\documentclass[pdflatex,sn-mathphys]{sn-jnl}

\jyear{2023}%
\usepackage{tabularx}
\usepackage[T1]{fontenc} 
\usepackage{verbatim} 
\theoremstyle{thmstyleone}%
\theoremstyle{thmstyletwo}%

\theoremstyle{thmstylethree}%

\begin{document}

\title[How to Integrate Digital Twin and Virtual Reality in Robotics Systems?]{How to Integrate Digital Twin and Virtual Reality in Robotics Systems?\\Design and Implementation for Providing Robotics Maintenance  Services in Data Centers}


\author*[1]{\fnm{Lin} \sur{Xie}}\email{lin.xie@utwente.nl}

\author[2]{\fnm{Hanyi} \sur{Li}}\email{hli@hanningzn.com}


\affil*[1]{\orgdiv{University of Twente}, \orgaddress{\street{Hallenweg 17}, \city{Enschede}, \postcode{7522NH}, \country{The Netherlands}}}

\affil[2]{\orgdiv{Beijing Hanning Tech Co., Ltd}
}



\abstract{In the context of Industry 4.0, the physical and digital worlds are closely connected, and robots are widely used to achieve system automation. Digital twin solutions have contributed significantly to the growth of Industry 4.0. Combining various technologies is a trend that aims to improve system performance. For example, digital twinning can be combined with virtual reality in automated systems. This paper proposes a new concept to articulate this combination, which has mainly been implemented in engineering research projects. However, there are currently no guidelines, plans, or concepts to articulate this combination. The concept will be implemented in data centers, which are crucial for enabling virtual tasks in our daily lives. Due to the COVID-19 pandemic, there has been a surge in demand for services such as e-commerce and videoconferencing. Regular maintenance is necessary to ensure uninterrupted and reliable services.  Manual maintenance strategies may not be sufficient to meet the current high demand, and innovative approaches are needed to address the problem. This paper presents a novel approach to data center maintenance: real-time monitoring by an autonomous robot. The robot is integrated with digital twins of assets and a virtual reality interface that allows human personnel to control it and respond to alarms. This methodology enables faster, more cost-effective, and higher quality data center maintenance. It has been validated in a real data centre and can be used for intelligent monitoring and management through joint data sources. The method has potential applications in other automated systems.}




\maketitle

\section{Introduction}
We are witnessing an accelerated development of new and emerging technologies (e.g. mobile technologies, virtual reality, digital twins) in the world of Industry 4.0, where the physical and digital worlds are closely linked. Today, thanks to advances in sensor technology, professional service robots are growing rapidly (by 131,800 units (+41\%) and US\$6.7 billion worldwide (+12\%) in 2020; see \cite{rotstatistics}). According to the International Organization for Standardization, a \textit{service robot} is defined as ``a robot that performs useful tasks for humans or equipment, excluding industrial automation applications'' (\cite{defrobot}). Unlike industrial robots, service robots can operate in unconstrained everyday environments. The International Federation of Robotics has published the top five applications for service robots in 2020, particularly autonomous mobile robots (see \cite{rotstatistics}; we will refer to them as \textit{mobile robots} in the rest of this paper). Most of them will operate in indoor environments for production and warehousing (see, for example, an overview of automated warehouses in \cite{azadeh2019robotized, xie2021introducing, xie2022formulating}). 

Automated systems are usually very complex. Before bringing such systems into reality, simulation is widely used as a virtual model to test the functionalities of the physical assets in such systems, e.g. to test a robot's navigation and localization algorithms (as in \cite{merschformann2017rawsim} for an automated warehouse), or to test different solution approaches for different problems in such systems to maximize performance (as in \cite{xie2021introducing} for an automated warehouse). The increasing development of sensors and the Internet of Things (IoT) allows the virtual model to interact with the physical system throughout its lifecycle. A digital twin (DT) is a dynamic and digital representation of a real object or system. Over the lifetime of the object, the digital representation is dynamically updated based on data provided by its physical counterpart. A more abstract definition of the DT can be found in \cite{van2021resilient}, while various definitions are listed in \cite{lim2020state}. Nowadays the DT is considered as a model to be used as a basis for simulations, analyses, etc. (\cite{dietz2020digital}). An example of a simplified DT for an automated warehouse can be found in \cite{xie2018simulation} and \cite{barosan2020development}. A DT of a system is expected to save time, money and effort by helping to identify system failures before they occur in the real system. DTs are widely used in industries such as product and manufacturing design, automotive (electric vehicles), predictive maintenance, healthcare, and so on. A recent review of various applications, especially in electric vehicle drive systems, can be found in \cite{ibrahim2022overview} and \cite{niaz2021autonomous}.

DTs have recently been used in automated systems by engineering researchers to test localization algorithms, robot motion (\cite{lumer2021towards}) and safety (\cite{almeaibed2021digital}), and the operating environment (e.g., warehouse layout options in \cite{stkaczek2021digital}) in a virtual environment before they are implemented in the real object. A recent overview paper on DT and its future perspectives can be found in \cite{lim2020state}, while a literature review on DT applications in manufacturing can be found in \cite{cimino2019review}. 

Although there is a trend for mobile robots to become increasingly autonomous (\cite{alatise2020review}), there are still some situations that require teleoperation by human personnel, e.g. in hazardous and dangerous environments where unexpected robot behavior must be avoided. \textit{Virtual Reality} (VR), thanks to its natural acting capabilities, is a well-adapted means to help a human interact with a robot remotely. VR is a representation of reality and it simulates a virtual environment that immerses the user to the extent that he has the feeling of being there (see \cite{wohlgenannt2020virtual}). Feedback from the environment is provided through images, sound, sensory stimuli, and interaction. More on the definition of VR and the use of VR in practice and research can be found in \cite{wohlgenannt2020virtual}. According to \cite{roldan2019multi}, VR is suitable when personnel are away from the mission, the mission is dangerous, the scenario is large, or personnel have to manage many resources (e.g. multi-robot systems). We need this technology especially when personnel and robots can interact with the real and virtual objects of the scenario. VR has been widely used in manufacturing, from design and training, programming and process monitoring to maintenance (\cite{oyekan2019effectiveness}). More recently, VR has been used to teleoperate mobile robots, e.g. in \cite{kot2018application} for remote military use and in \cite{li2022towards} for inaccessible remote environments, while there is an overview of applications in engineering research projects in \cite{wonsick2020systematic}, e.g. on controlling and interacting with robots (excluding mobile robots). 

VR provides a safe and cost-effective environment for testing concepts and hypotheses, such as a robot arm, before deployment in the real world. In addition, the feedback, knowledge, and optimized virtual models obtained in a VR environment can be transferred to the real-world object. To achieve a seamless transfer, a DT of the real-world environment, including the workshop and the robot arm, is required. 

\subsection{Contribution and structure of the paper}
The combination of DT and VR technologies has recently been involved in engineering publications, mainly on the design of human-robot collaboration in an industrial manufacturing process (\cite{oyekan2019effectiveness, havard2019digital, perez2020digital}), the methods for programming and controlling robots (\cite{burghardt2020programming, kuts2019digital}), a laboratory application in a circular economy (\cite{rocca2020integrating}), the calibration of mobile robots (\cite{williams2020augmented}), or safety training (\cite{kaarlela2020digital}). These have been isolated efforts with no guiding principles, plans, or concepts to clarify the combined use. This is the gap we are addressing. 

The contributions of this work can be summarized as follows:
\begin{enumerate}
    \item We propose a novel concept to combine different technological trends, including digital twinning, VR and mobile robots (the theoretical result), to perform monitoring in DCs.
    \item The concept is implemented and validated in a novel case study in a real data center to perform monitoring and inspection (the practical result). Such an automated maintenance system in data centers has not been considered in the literature so far (see more details in section~\ref{sec:background}).
\end{enumerate}
The validation shows that the use of the DT-based methodology is feasible. Thanks to the integration of VR, the proposed approach is more visible; furthermore, it reduces costs and provides higher quality at high speed, as proven in real operations. Therefore, it is very likely that potential applications of our method will be found in different areas, such as other asset-centered organizations like a power plant. 

The paper is organized as follows. We begin with a description of the motivation and reasons for choosing data centers as our case study in section~\ref{sec:background}. Then, our proposed concept and its development are described in sections~\ref{sec:concept} and \ref{sec:develop}, respectively, while the results of the application in a real data center are presented in section~\ref{sec:validation}. We will conclude this work with a short summary and an outlook in section~\ref{sec:concl} with some suggestions for future research.

\section{Background of maintenance in data centers}\label{sec:background}
In this section, we describe the importance of performing maintenance in data centers in subsection~\ref{subsec:DC}, common maintenance strategies in subsection~\ref{subsec:tramet}, and innovative approaches in subsection~\ref{subsec:inno}.

\subsection{Data center} \label{subsec:DC}
A \textit{data center} (DC) containing a large number of servers and computers can provide internet services for many companies in the world, such as Google and Amazon. Servers and computers are stored in racks (e.g. 42U 1200mm deep racks from Lenovo) in a DC. According to the categories mentioned in \cite{bieser2018role}, a DC can be an in-house DC (hardware and infrastructure are privately owned), a co-locator DC (e.g., modular service packages are offered to users with or without information technology (IT) applications), or a cloud DC (services are offered over the Internet without users owning the IT infrastructure). Each DC maintains important environmental factors such as proper temperature and humidity. Due to the high demand for a wide range of services (such as web search, e-commerce, video streaming, gaming, high-performance computing, and data analytics), the number of DCs is growing worldwide; in particular, the number of cloud DCs (such as Amazon EC2 and Microsoft Azure) has grown rapidly recently. The COVID-19 pandemic has significantly increased the use of online tools for applications such as video conferencing, virtual business events, and remote education. This shift has been enabled by cloud environments, especially public clouds. There are more than 7 million DCs worldwide. According to Synergy Research Group, in addition to the 600 operational hyperscale data centers worldwide, another 219 are in the planning or construction phase (\cite{datacenter}). The largest technology companies (Google, Facebook, Amazon, and Microsoft) have more than 50 DCs with more than 1.3 million servers.

The primary mission of a DC is to provide reliable and continuous service. Downtime in DCs should be avoided because it has a massive negative impact on the business, such as loss of capacity, cost of defects, cost of delays, and also the indirect costs caused by, for example, reputational damage. According to research conducted by the Ponemon Institute in 2016, the average cost of DC downtime was approximately $\$$8,000 per minute, and due to the average reported incident length of 95 minutes in the survey, the average cost of a single downtime event was $\$$740,357, a net increase of 38\% compared to 2010 (\cite{ponemon}). According to \textit{Fortune}, Facebook lost $\$$99.75 million in revenue during the October 7, 2021 outage that lasted more than six hours. 
In literature, limited researches concentrate on failure analysis in DCs, e.g. \cite{gill2011understanding} for network failures, \cite{guenter2011managing} for hardware failures and \cite{ram2014modeling} for various failures, such as failure of redundant server, switch, router or server cooling. There are some methods implemented in the literature to reduce failures. For example, high temperatures and repeated on-off cycles of a server can increase hardware failures. To avoid them, the workload distribution of servers is optimized by considering thermal characteristics (\cite{wang2009towards, guenter2011managing}) or with balanced energy consumption and quality of service (\cite{bayati2016managing}), or some new cooling systems are introduced (\cite{chen2013optimization, cho2015development, li2016data}).
\subsection{Traditional maintenance strategies in DCs} \label{subsec:tramet}
As mentioned in \cite{ignore}, 30\% to 40\% of system timeouts due to infrastructure hardware failures can be avoided through regular \textit{preventive maintenance}. Note that IT infrastructure includes servers, switches, routers, power supplies, and cooling and monitoring systems. Preventive maintenance is defined as ``maintenance carried out at predetermined intervals or according to prescribed criteria and intended to reduce the probability of failure or delegation of the function of an item'' (DIN EN 13306). Preventive maintenance can include replacing backup batteries or generators, replacing servers and switches, upgrading software, etc. In addition, preventive maintenance is either performed according to an established schedule without prior condition surveys (\textit{scheduled}) or scheduled on demand or on the basis of performance and parameter monitoring (\textit{condition-based}). More about the definition and categories of maintenance can be found in \cite{bieser2019assessing}. The same literature provides an overview of current maintenance concepts. The importance of maintenance in DCs is highlighted in \cite{bieser2018assessing} from an operational and economic perspective, and a preventive maintenance approach is introduced. An overview of the main practical issues of DCs in terms of maintenance management can be found in \cite{abadi2020data}. 

In this work, we focus on condition-based maintenance based on regular monitoring. System \textit{monitoring} is one of the research topics in maintenance management. The traditional methods of monitoring a DC include using a monitoring system (with sensors/cameras installed at fixed locations) and manual inspection. Using the first method, it is expensive to install enough sensors/cameras to monitor all assets in a DC. In addition, it is impossible to confirm the situation on site when failures occur, and human personnel are required to be on site. The second method requires human staff to check the status of all assets, which is very time consuming. In addition, the limitations of human vision and skill lead to most technical errors in monitoring. Each infrastructure in a DC is monitored by the monitoring system; when failures occur, the system detects them and alerts the human staff. The traditional method of monitoring a DC requires human staff to be on site.
 Wireless monitoring is discussed in \cite{kadir2015wireless}; a 3D gaming platform is developed to enable real-time monitoring and display visual analysis of the operating condition in a DC (\cite{hubbell2015big}).

As mentioned in \cite{abadi2020data}, innovative approaches are needed to expand the scope of traditional maintenance strategies in a DC. A recent overview in \cite{levy2021emerging} mentions three main technological trends in DC management automation that should receive more attention in the near future: 1) intelligent monitoring and management systems using data science (e.g., predictive maintenance in \cite{decker2019big}); 2) simulation tools incorporating artificial intelligence and digital twinning; and 3) robotics for process automation. The second and third technological trends will be the focus of the remainder of this work, as they enable the automated collection of vast amounts of sensory information, which is seen as the basis for predictive maintenance. 

\subsection{Innovative maintenance approaches in DCs}\label{subsec:inno}

With the increasing size of DCs, monitoring in DCs is very challenging and automation is a mandatory requirement of scale. Monitoring is a repetitive task, and autonomous mobile robots can be used to perform round-the-clock monitoring to increase process automation and productivity, and reduce human error. In addition, the environment in DCs is not human friendly - it is noisy, it can be cold and it can be hot. There is a growing demand for hyperscale DCs that are more like warehouses, where robots need to navigate to specific locations to perform tasks. It is also more cost-effective to use mobile robots with built-in sensors to perform monitoring than to install many sensors at fixed locations in DCs. Note that mobile robots are used to perform monitoring and inspection in various industrial applications, such as offshore platforms (\cite{gehring2021anymal}), power transmission lines (\cite{alhassan2020power}), oil and gas industry (\cite{yu2019inspection, roh2001activ}), and deep underground mines (\cite{zimroz2019should}). An overview of monitoring and inspection with mobile robots can be found in \cite{kroll2008survey} and \cite{katrasnik2009survey}. 

Recent research in DC monitoring using mobile robots focuses on a robotic platform for \textit{environmental monitoring} (see \cite{rosa2014towards, russo2016novel, terrissa2019robotics}). Environmental monitoring aims at measuring the environmental qualities in DCs for diagnostic and control purposes. The sensors are installed in mobile robots to monitor temperature, humidity and air quality. The sensors can be, for example, a thermal camera and temperature/humidity sensors (\cite{rosa2014towards,russo2016novel}). In this work, we focus not only on environmental monitoring, but also on \textit{asset monitoring} in DCs. Asset monitoring aims to check the quality of assets in DCs around the clock, including servers, computers, switches, routers, network equipment, and so on. For such repetitive and time-consuming tasks (due to the large number of assets in DCs), mobile robots are required to perform inspections. In this way, asset failures can be quickly detected (even predicted) and it is possible to ensure continuous services from a DC. However, such asset monitoring using mobile robots in DCs has not been considered in the literature. 

DCs are asset-centric organizations, and it is important to avoid significant asset downtime (as mentioned above). A DT plays an important role in this, as it is the virtual counterpart of an asset, allowing organizations to digitally mirror and manage an asset throughout its lifecycle. To virtually represent the life and behavior of the asset, a DT must incorporate all types of data from multiple domains related to the asset and continuously provide information about the condition of the asset, including its status and maintenance needs. Citigroup is an example of how it integrated the digital twin of the DC into its workflow to improve the efficiency of its New York City DC. DT-designed solutions quickly reduced inlet temperatures by 27\% and generated over \$290,000 in potential annual energy savings (see \cite{Citigroup}). 3D reconstruction methods can provide the DT component to drive a vision-guided robotic inspection system in the manufacturing and remanufacturing industries (\cite{khan2021inspection}). However, the DC digital twin has not yet been addressed in the literature. 

Asset monitoring is complex, and human intervention is still needed to deal with robot errors, to recheck some assets, and to replace assets. Note that robots that can automatically replace faulty disks are still being developed, such as Alibaba's Tianxun robot \cite{RobotsInDCs2, RobotsInDCs1}. Therefore, there is a strong need for humans to interact with robots and their services in the DT interface, e.g., to control robots to re-check a faulty disk. The DT lacks human interaction capabilities. 

In this work, we focus on the use of VR technology in a digital twin of a DC to enable human interaction with robots to perform monitoring (i.e., automatic inspection, faulty disk location, disk replacement), which is rarely addressed in the literature. One existing paper is about an educational VR application for teaching cybersecurity concepts in a DC (\cite{seo2019using}). Furthermore, interaction with robots also allows remote control of robots in case of failures.
\section{The proposed concept} \label{sec:concept}
\begin{figure}[t]
    \includegraphics[width=\textwidth]{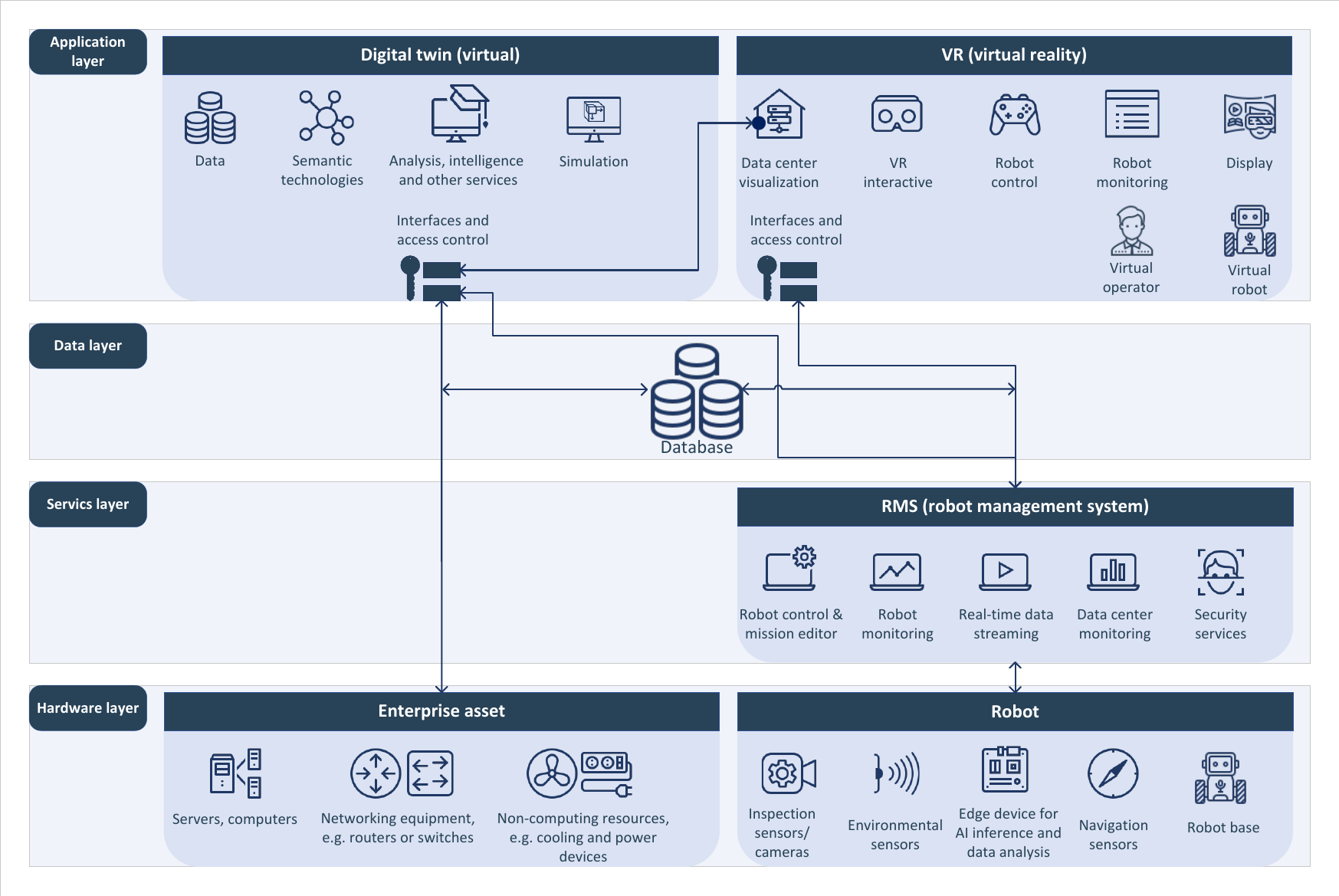}
    \caption{The concept for the realization of the digital twin and virtual reality to provide robotics services in a DC.}
    \label{fig:concept}
\end{figure}

In this work, we propose a new concept to combine DT and VR to provide robotic services (see Figure~\ref{fig:concept}). Here, we use the DC environment and assets as well as monitoring services to better understand the concept. They can be applied to other use cases, such as robotics services in smart framing or other asset-centric organizations, by changing the assets and the content of the services. This concept extends the DT paradigm proposed by \cite{dietz2020digital} with different layers, while VR is combined to provide visualization of DT objects and remote control of the robots. 

Four layers are involved in our concept: the \textit{hardware layer} (physical layer), which contains enterprise assets and a mobile robot (including a robot base, an edge device, and various installed sensors, etc.); the \textit{service layer} for providing different types of robotics services in a robot management system (RMS) based on the information collected by the robot, e. g. the \textit{data layer} for providing/collecting data (of any type, e.g. sensor-based, reports, video data) to/from different layers; and the \textit{application layer} (digital layer) for representing the virtual counterparts to the assets and the robot in the hardware layer as well as their displays. 

\begin{figure}[h]
    \includegraphics[width=\textwidth]{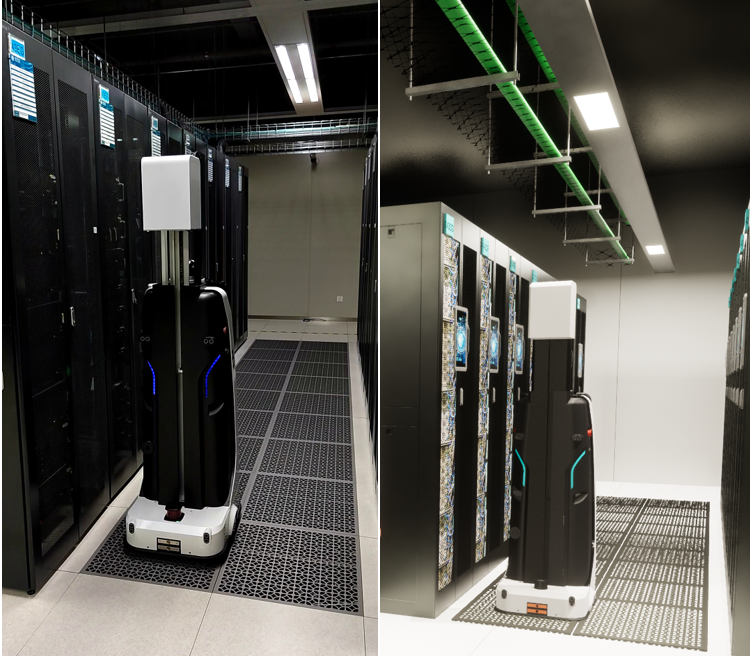}
    \caption{Real DC vs Digital Twin DC (including an inspection robot).}
    \label{fig:real_virtual}
\end{figure}

In the application layer, a virtual DC is modeled by the DT and the information provided by the service layer is displayed by the VR. Figure~\ref{fig:real_virtual} shows a real DC and our developed DT DC in the same scene.
Our DT is built from the following blocks, similar to \cite{dietz2020digital}:
\begin{itemize}
    \item \textit{Data} includes static and dynamic data. The static data is provided by the assets from the hardware layer to model the DC, which is considered to be static, while the dynamic data, including the sensor/camera-based inspection data (e.g. temperature, humidity, server status) and robot information data (e.g. robot status), is received from the RMS. 
    \item \textit{Semantic technologies} are used to describe the relationships between data elements to better understand their meaning.
    \item \textit{3D simulation} is applied and combined with data to represent a DC to study its behavior. 
    \item \textit{Analysis, intelligence and other services} are supported in the DT ranging from asset monitoring to autonomous driving of robots. For example, historical data can be analyzed to perform predictive maintenance. 
    \item \textit{Interfaces and access control} are the mechanisms that connect the virtual and real worlds. They allow data exchange and synchronization. The bi-directional connection is enabled between the asset/robot and its twin, also between the DT and VR. The development of the interface is described in section~\ref{sec:develop}.
\end{itemize}

Sherman \& Craig defined four key elements of the VR experience in \cite{VR2018book}: virtual world, immersion, sensory feedback, and interaction. VR is seen as one of the tools with the most potential to enable human-robot collaboration. The most commonly used VR headsets connected to a computer are Oculus Rift, HTC Vive and Pico Neo. Our DT system includes several interactive VR elements. Our VR is built from the following blocks.
\begin{figure}[h]
    \includegraphics[width=\textwidth]{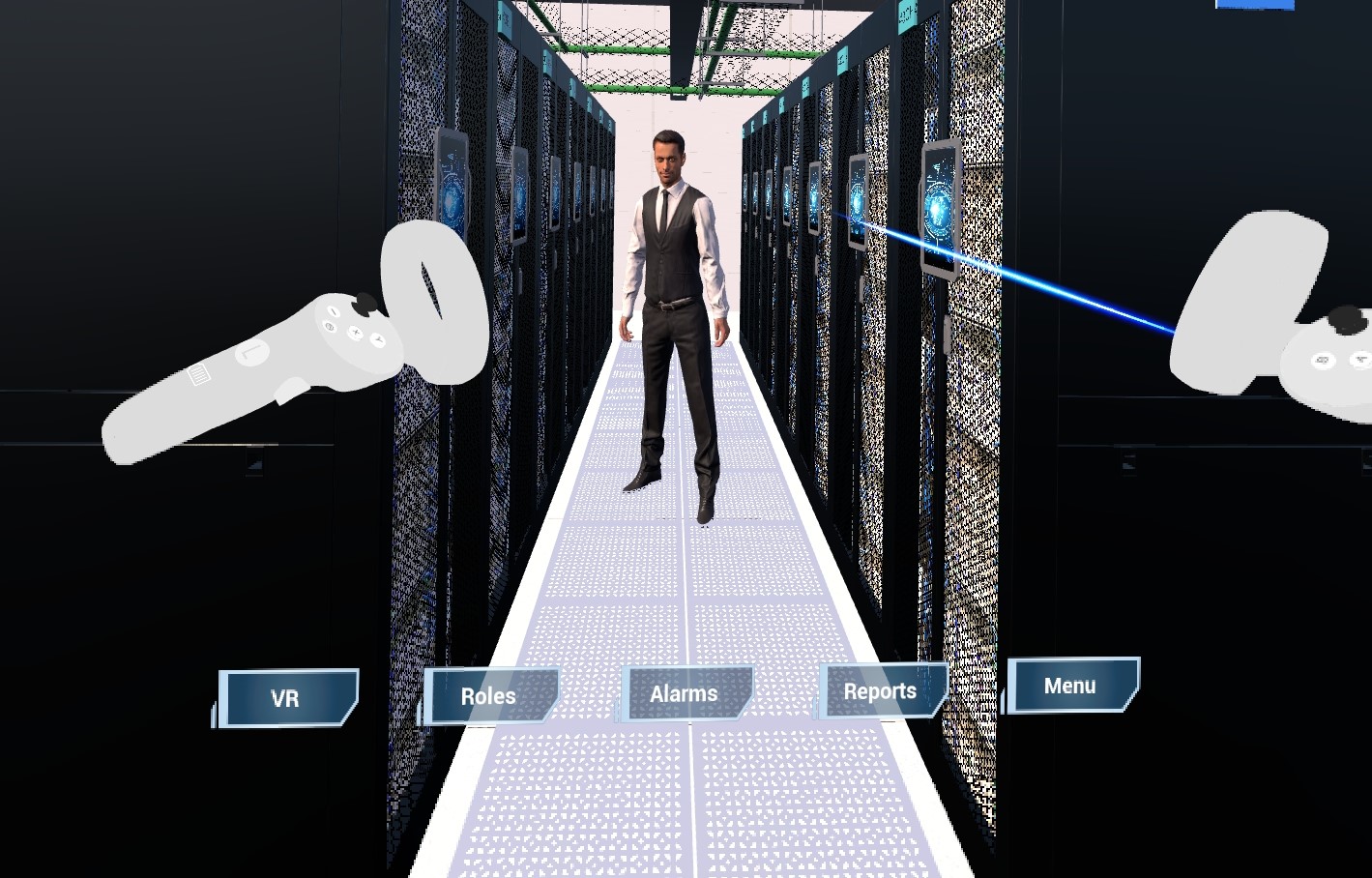}
    \caption{The VR data center with virtual staff from the robot's perspective.}
    \label{fig:vr_data_center}
\end{figure}

\begin{figure}[h]
    \includegraphics[width=\textwidth]{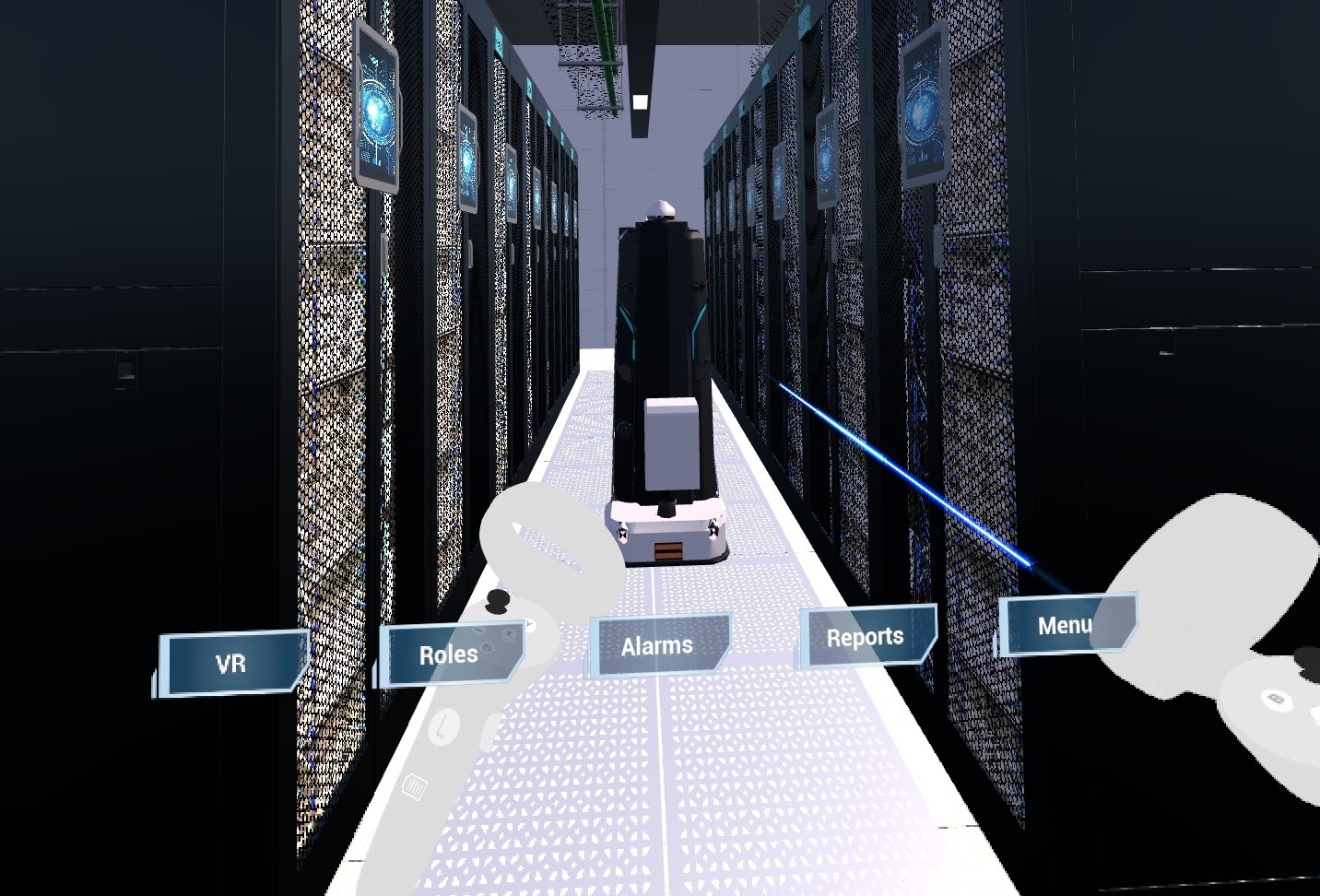}
    \caption{The VR data center with a virtual inspection robot from the virtual staff's perspective.}
    \label{fig:vr_data_center_staff_view}
\end{figure}

\begin{itemize}
    \item \textit{DC visualization} shows the virtual DC environment which is reconstructed in the DT.
    \item \textit{VR interactive} includes teleoperation and visualizes the robot together with the DC environment in 3D, information such as the state of the robot and knowledge database.
    \item \textit{Robot control} sends a request for the robot control service from the RMS to fulfill a given task.
    \item \textit{Robot monitoring} sends a request for the robot monitoring service from the RMS, including a representation of the robot and its movements, information about the robot's state and the measurements from on-board sensors, images from cameras, and the progress of the ongoing inspection tour.
    \item \textit{Display} includes the display of asset information, video stream, inspection reports, alarms and visitor information. 
    \item \textit{Interfaces and access control} have the same purpose as in the DT. The bi-directional connection is enabled between VR and the RMS, and also between the DT and VR. There is more about the development of this interface in Section~\ref{sec:develop}.
    \item \textit{Virtual staff} is a realistic human character, which is developed by MetaHuman Creator, a free cloud-based app for creating a fully rigged, photorealistic, digital human (see \cite{MetaHuman}). 
\end{itemize}

Unlike the usual VR interface for a single robot as in \cite{roldan2019multi}, where the interface is centered in the robot and its payload, and a first-person view of the robot is adopted, we have different views in our VR interface, including a first-person view of the robot in Figure~\ref{fig:vr_data_center} and a first-person view of the virtual staff as shown in Figure~\ref{fig:vr_data_center_staff_view}. Using the images from the robot's camera in our virtual environment, we can identify an asset and display its type, maintenance manual, etc. via the interface, display a real-time video stream, identify the location of an alarm, or identify the person (visitor or employee, etc.) using facial recognition (for security purposes) and display their personal information. We can also display the inspection reports through the VR interface. 
Our VR interface can easily support multi-robot systems by selecting the camera of one of the robots and we can change the view from one camera to another. As shown in both figures, a virtual interactable object is attached to the front of each rack, which has the following two functions. First, it provides users with inspection information about their rack. Second, it can be used as an input device within VR, where the user can give commands to the system by interacting with warning messages.

\begin{figure}[h]
    \includegraphics[width=\textwidth]{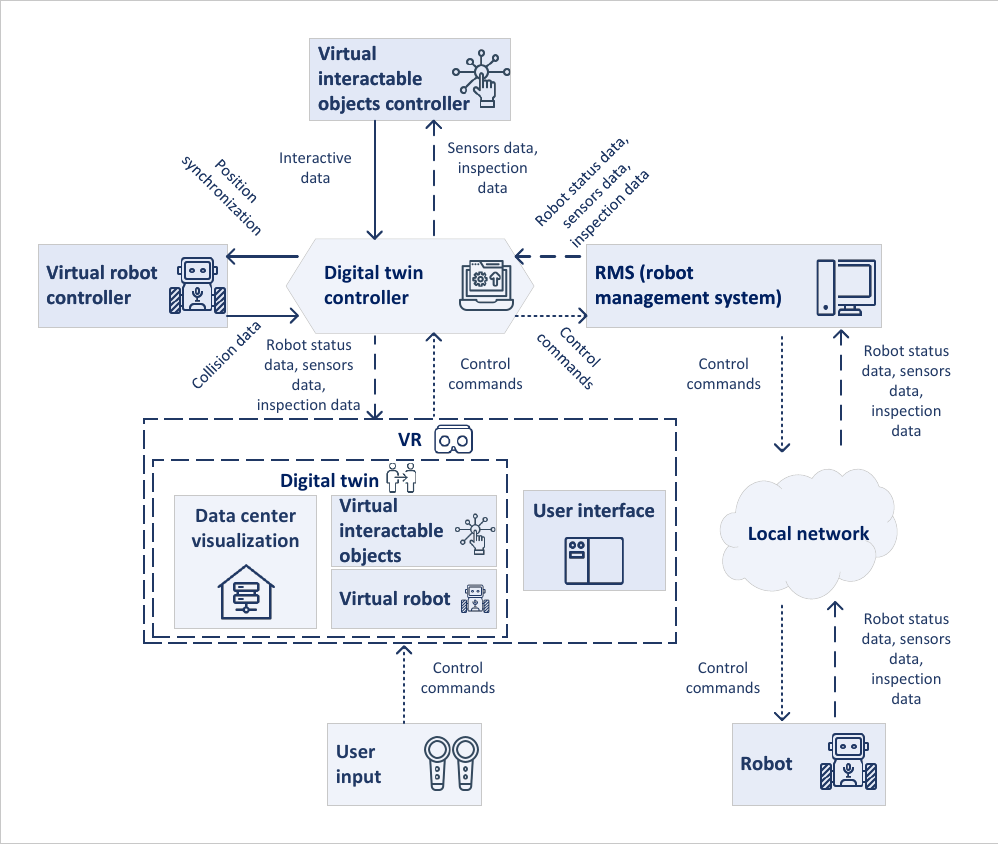}
    \caption{DT and VR system diagram (expanded and modified on \cite{Kuts2019dt}).}
    \label{fig:digital_twin_sync}
\end{figure}

 In~Figure~\ref{fig:digital_twin_sync}, we extend and modify the schematic representation of the DT and VR system from \cite{Kuts2019dt} designed for industrial robot arms. Note that such a system is widely used to program motions for an industrial robot arm. Similarly, we use the DT controller as the core, and we import the computer-aided design (CAD) model of the robot (from the robot manufacturer) into the VR environment. The virtual robot receives position synchronization from the DT controller and sends collision data. The real robot sends status data and receives control commands through the local network. The following modifications are made in our system. First, we connect the DT controller to the RMS (instead of a robot controller) to enable services such as receiving robot status and inspection data. Second, in addition to the virtual robot, we have a more complex virtual environment that also includes a virtual DC and virtual interactable objects. The virtual interactable object controller sends interaction data to the DT controller and receives inspection data from it. When the user sends commands, they are sent through VR to the DT controller. The DT controller then sends commands to the RMS, and the RMS controls the robot over the local network. After the real robot has sent the status and inspection data to the RMS via the local network, the data is sent to the DT controller. The actual data is displayed in VR, and the inspection data is updated in the virtual tablet by the virtual interactable object controller.

\section{Development of our concept}\label{sec:develop}
\begin{figure}[h]
    \includegraphics[width=\textwidth]{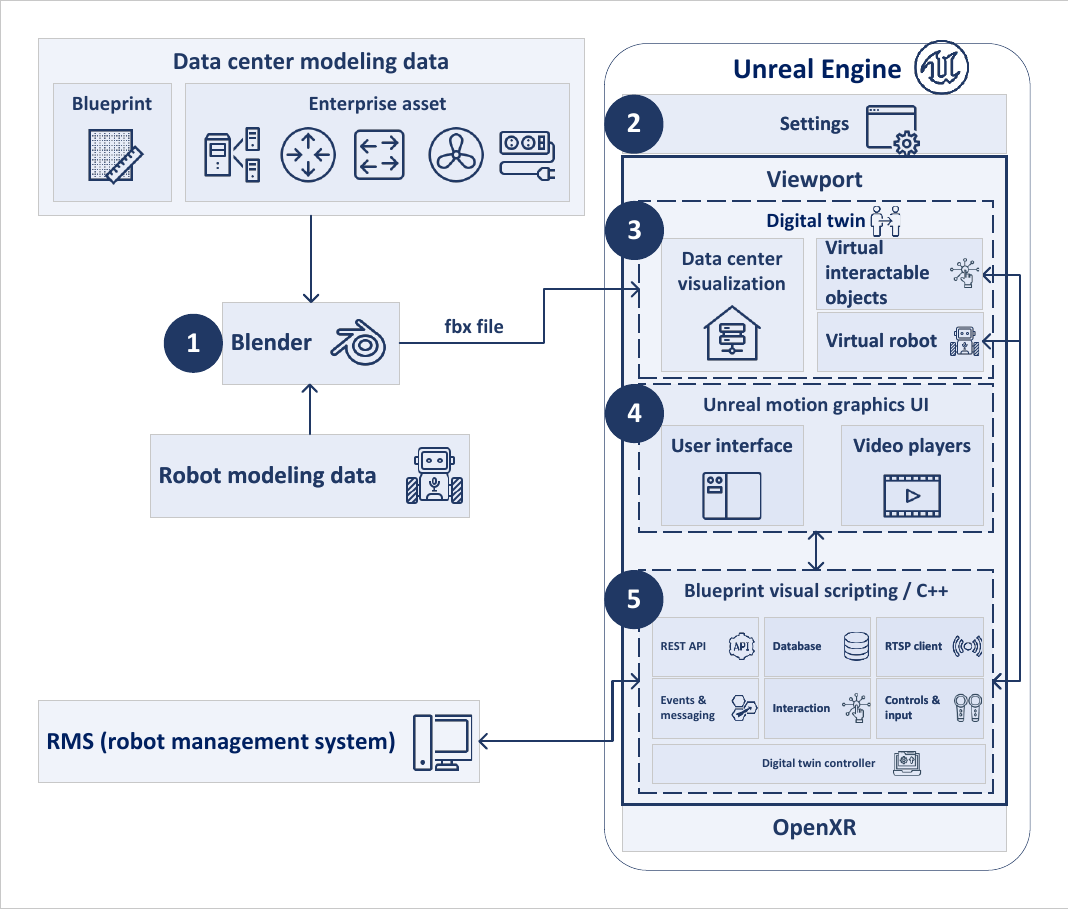}
    \caption{The development process for our DT and VR.}
    \label{fig:dt_vs}
\end{figure}

To implement the concept explained in the previous section, we designed a development process as shown in Figure~\ref{fig:dt_vs}. Our VR interface is developed based on the use of \textit{Unreal Engine} and \textit{OpenXR}. Unreal Engine is a 3D computer graphics game engine developed by Epic Games. The authors in \cite{hubbell2015big} list several reasons why a game engine can be successfully applied to DC infrastructure management, including the ability to convey interactive, actionable data, platform independence, ease of use for users, and the use of tools in the 3D environment, such as colors, shapes, morphing an object, and in-game sounds. OpenXR is a royalty-free, open source standard that provides high-performance access to XR platforms and devices (see \cite{OpenXR}). The OpenXR plugin is a library of built-in classes and components to support VR.   

In the first step of the development process in Figure~\ref{fig:dt_vs}, we collect the data center modeling data, such as a blueprint of a DC and the enterprise assets. 
We use \textit{Blender}, which is a free and open source 3D animation suite, to create the DC model, while the robot modeling data is used to create the robot model. Then we export the model into the .fbx file format (a format used to exchange 3D geometry and animation data), which is used as the import chain of Unreal Engine.  


\begin{figure}[h]
    \includegraphics[width=\textwidth]{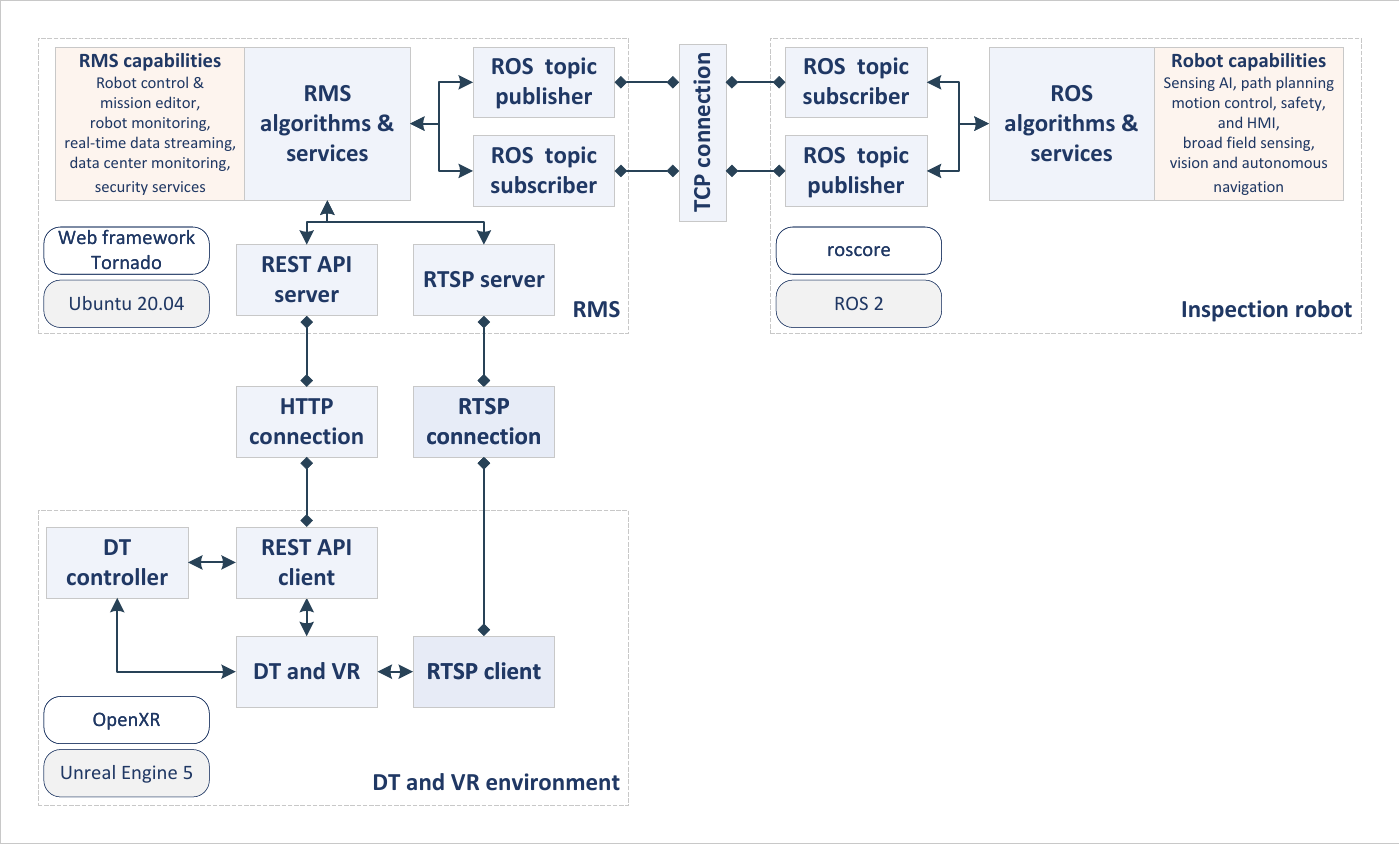}
    \caption{Overview of the system technical architecture.}
    \label{fig:technical_architecture}
\end{figure}

In the second step, we determine the settings in Unreal Engine, including the settings for Unreal's built-in VR support, device-specific toolkits (for example: SteamVR plugin, Oculus integration toolkit, Pico Unreal XR SDK), application toolkits, target platform, and XR SDK. In the third step, we place the DT in the scene to visualize the DT, the robot, and interaction objects such as a virtual tablet, as shown in Figure~\ref{fig:real_virtual}. In the fourth step, we set up the user interface and video player in the Unreal motion graphics UI (the area where all UI elements should be located), while in the fifth and final step, we write the blueprint visual scripting scripts or C++, which includes the following main functionalities REST API handler; database handler; RTSP (Real-Time Streaming Protocol) client for receiving the video stream; event and messaging; interaction (e.g., interaction with the virtual tablet, virtual robot, etc.); and controls and input for controlling the robot. More technical details of these functionalities can be found in Figure~\ref{fig:technical_architecture}. 



As shown in Figure~\ref{fig:technical_architecture}, the basic components and structures of our DT and VR interaction system are an inspection robot, the RMS, and the DT and VR environment. 

The Robot Operating System (ROS) is the de facto standard for robotic software development and provides a collection of packages and tools for developing distributed robotic applications. The inspection robot developed using the ROS algorithms and services has capabilities such as AI for sensing, path planning, motion control, safety and human-robot interaction (HRI), wide field sensing, vision, and autonomous navigation. 

The robot and the RMS are connected via a Transmission Control Protocol (TCP) connection over a local network using the ROS TCP connection. The RMS is developed in Python. We use the web framework and the asynchronous networking library Tornado as the back-end on Ubuntu and MySQL-based data storage, while the front-end is developed in JavaScript using Vue.js (an open-source model-view-view-model front-end JavaScript framework). To enable the RMS to communicate with robots from different manufacturers, it is extended by other communication interfaces, e.g. MQTT. The RMS has capabilities such as robot control and mission editor, robot monitoring, real-time data streaming, data center monitoring, and security services, which are implemented as RMS algorithms and services. 

The RMS provides REST API and RTSP services to our DT and VR clients. The RMS and our DT and VR clients are connected via HTTP (Hypertext Transfer Protocol) as a REST API server and client, respectively, while the RTSP ("real-time streaming protocol") connection allows the RTSP server and client to be connected over a local network. The DT and VR client runs on the OpenXR backend, and its DT controller, as described in Figure~\ref{fig:digital_twin_sync}, is used to connect the RMS to VR and DT components such as the virtual robot. The DT and VR client sends commands and receives data from the RMS via the RESP API client, while the video stream data is received by the RTSP client.

In order to achieve all of the above, the following special plugins or APIs of Unreal Engine 5 are required for development in the DT and VR environment. 
\begin{itemize}
    \item The \textit{Multi Task 2 plugin} is used to enable multi-threading in Blueprints Visual Scripting System (\cite{MultiTask2}).
    \item The \textit{VLC plugin} (VLC is a free and open source cross-platform multimedia player and framework; see \cite{VlcMedia}) is selected, which is adapted and built for Unreal Engine 5. It is used to handle and play the RTSP stream.
    \item The \textit{VaRest plugin} is used to handle the REST communication (\cite{VaRest}). The DT controller receives the robot's position information, its motion information, its angle, the position for the vertical lifting/lowering module of the HD and thermal camera on the robot, and a number of related robot information from the RMS via the REST API every 0.5 seconds. This information needs to be synchronized with the DT of the virtual robot (see figure~\ref{fig:real_virtual}) and then visualized, which requires some smoothing on 3D graphics.
    \item The \textit{DataRegistries plugin} is used to store, combine, read and manage data from different sources for the implementation of the database handler (see \cite{DataRegistries}).    
    \item Mathematical APIs on the Unreal Engine are used to perform some interpolation operations to smoothly realize the position of the virtual robot (see \cite{MathInterpolation}).
\end{itemize}    




\section{Validation of our concept in a real DC}\label{sec:validation}
In this section, we first describe the real DC environment in Subsection~\ref{subsec:real_dc}, the robotics services in Subsection~\ref{subsec:real_rot_service}, and the data processing from various sources in Subsection~\ref{subsec:real_data}. We then describe the DC inspection workflow in Subsection~\ref{subsec:real_workflow} and the results of applying robotics services to monitor the real DC in Subsection~\ref{subsec:results}. 
\subsection{DC environment} \label{subsec:real_dc}
As a case study, we implement our concept in a data center with an area of 522.5 m$^2$, including 140 racks. It includes the following assets in the physical layer of Figure~\ref{fig:concept}:
    \begin{itemize}
        \item \textit{Computing resources} including servers, computers and networking equipment (servers and computers are located in racks).
        \item \textit{Non-computing resources} including cooling and power devices.
    \end{itemize}

An inspection robot (in the physical layer), as illustrated in Figure~\ref{fig:robot}, is used in the experiment. The robot includes the following two types of components:
\begin{figure}[h]
    \includegraphics[width=0.7\textwidth]{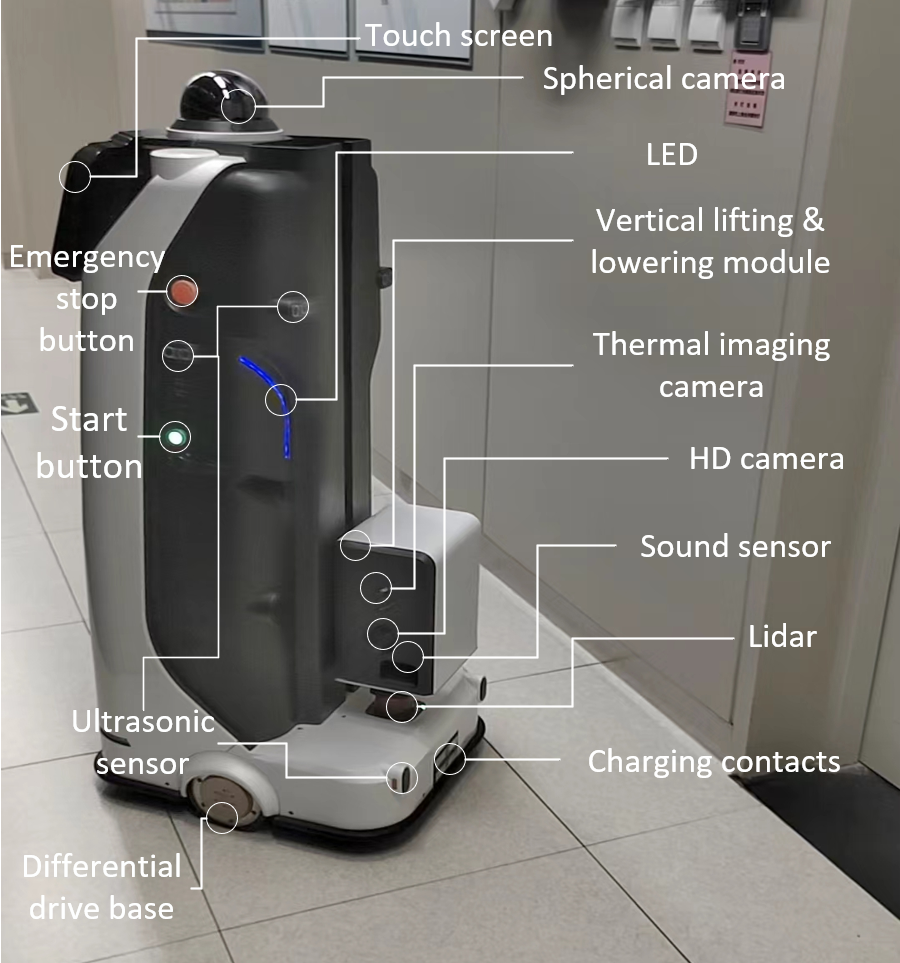}
    \caption{An inspection robot in a DC environment. }
    \label{fig:robot}
\end{figure}
\begin{itemize}
    \item \textbf{Components to inspect}
    \begin{itemize}
    \item \textit{Vertical lifting \& lowering module} to move the HD and thermal camera for changing inspection targets at different heights of the same location (max inspection height is 2.4 meters; see e.g. the robot in Figure~\ref{fig:real_virtual} in the lifting module). 
    \item \textit{Inspection sensors/cameras} including HD camera and thermal imaging camera used to collect equipment information and prepare for asset monitoring. The location of the server is predefined on the map based on the server number, so the camera can quickly locate the server number during the inspection;
    \item \textit{Environmental sensors} (not visible in the figure), including a temperature and humidity sensor, a noise detector, and an air quality sensor (PM 2.5 sensor), which are required to collect environmental information for DC environmental monitoring. 
    \item \textit{Touch screen, spherical camera, microphone and speaker} for Human-Robot Interaction (HRI) and video intercom function for remote call.
    \item \textit{Edge device} (not visible in the figure) for AI inference and data analysis for inspection (visual inspection tasks, acoustic and thermal inspection). It is able to detect the defect information from the light signals coming from the servers and extract its information with the help of a machine learning algorithm.
    \end{itemize}
    \item \textbf{Autonomous driving components}
    \begin{itemize}
    \item \textit{Built-in computer} (not visible in the figure) as the navigation PC, which can perform real-time and high-performance computation for sensing and localization, mapping, path planning and following, and mission planning.
    \item \textit{Navigation sensors} including Lidar and depth cameras (to determine the robot's position relative to the known environment, simultaneous localization and mapping (SLAM) by fusion of Lidar and depth cameras) and ultrasonic sensors (to detect the motion of objects and measure the distance to them), which are needed to enable the robot to navigate and avoid obstacles and collisions in a DC. For optimal obstacle avoidance, the depth camera can be installed; by using the depth camera, the robot can "see" and understand its surroundings. The localization process determines the robot's current position on the map based on input from the motor encoders, the inertial measurement unit (IMU) and the lidar.
    \item \textit{Differential drive base} to ensure the differential drive of the mobile robot, i.e. the movement of the robot is based on two separately driven wheels placed on either side of the robot body (the maximum speed of the robot is 1.5 m/s and the acceleration is limited to 0.3 m/s$^{2}$).
    \item \textit{Charging contacts} to contact the charging station for battery charging.
    \item The \textit{battery} (not visible in the figure) on this inspection robot has a capacity of 1.56 kWh (30.4 Ah at 51.2 V). The battery type is lithium-ion (Li-ion). The active operating time at full load is approximately 6 hours. It takes about 1 hour to fully charge (from 40\% to 100\%).
    \item \textit{Others}: The robot has an emergency stop button on each side and a start button; the wireless network antenna is located inside the robot; LED lights are used to display the moving or operating status of the robot and some execution status items such as charging and internal errors in the robot. A remote module (network relay module) sends a signal via TCP/IP to the door opener controller to ensure the opening and closing of the DC door.
\end{itemize}
\end{itemize}

\subsection{Robotics services in the DC} \label{subsec:real_rot_service}
    
\begin{figure}[h]
    \includegraphics[width=\textwidth]{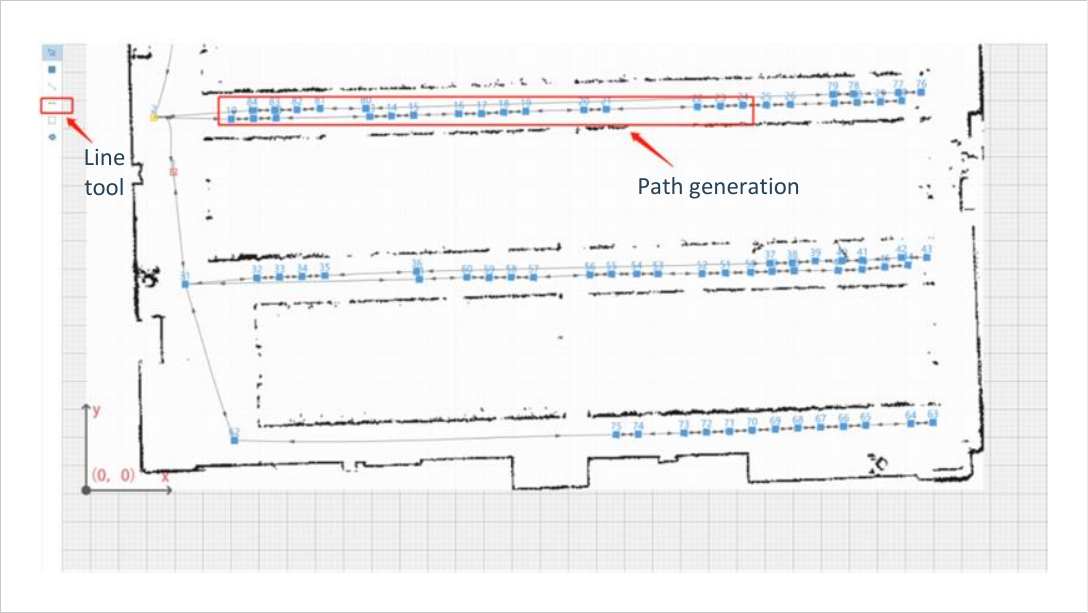}
    \caption{The UI of the RMS for mission planning.}
    \label{fig:task_plan}
\end{figure}

In the service layer of Figure~\ref{fig:concept}, the RMS receives the information collected by the robot and provides the following services via an a user-friendly UI (user interface) as shown in Figure~\ref{fig:task_plan}, which are accessible from the data and application layers. 

\begin{figure}[h]
    \includegraphics[width=\textwidth]{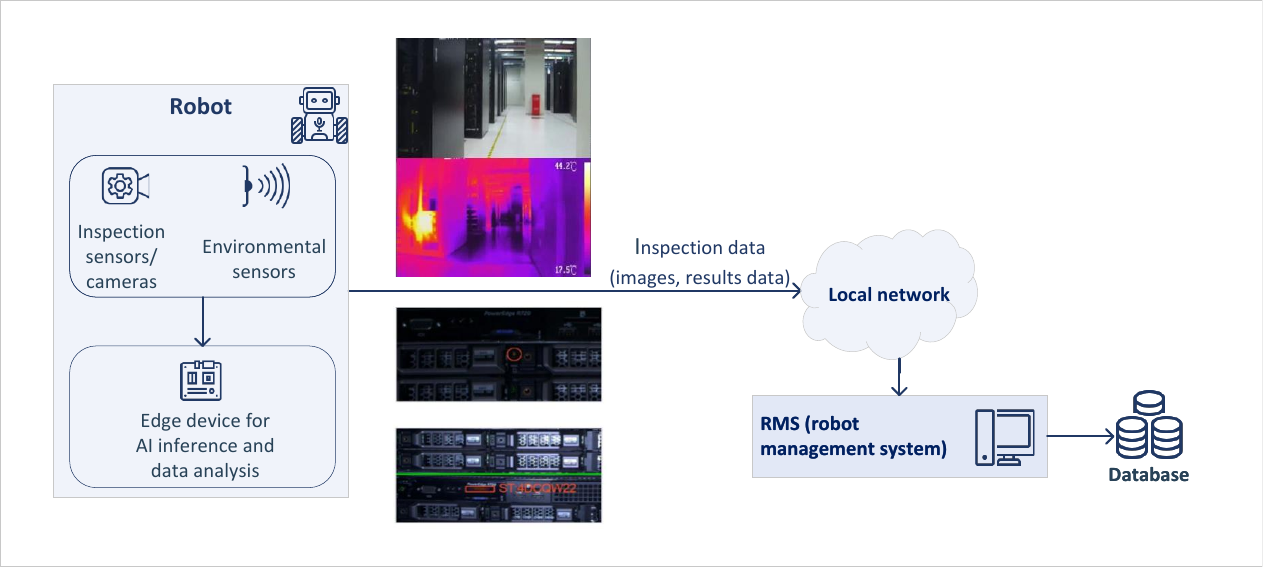}
    \caption{The data flow process of the DC inspection with robot. }
    \label{fig:ai}
\end{figure}
\begin{figure}[h]
    \includegraphics[width=\textwidth]{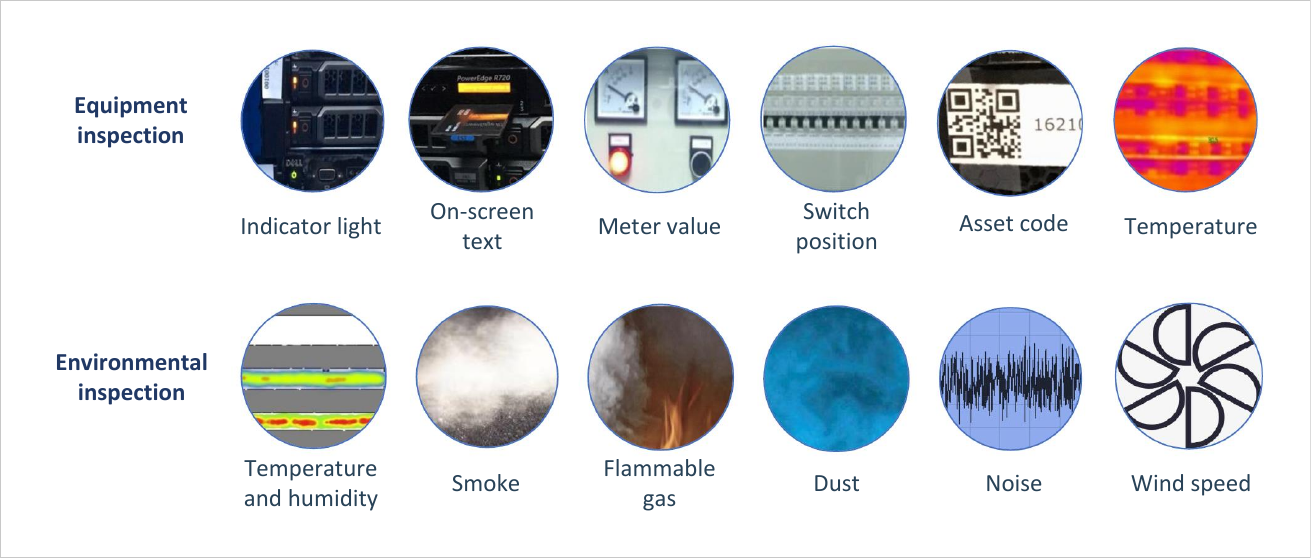}
    \caption{Equipment and environmental inspection.}
    \label{fig:ee_inspection}
\end{figure}

The following robotics services provided in the RMS are developed in the DC:
\begin{itemize}
    \item \textit{Robot control and mission editor}, which specifies the robot's tasks, including, for example, recharging and executing an inspection plan based on a mission plan. In Figure~\ref{fig:task_plan}, the path and inspection points (blue points with numbers) of the inspection mission are defined by the user. The user can select the mission and its start time. The RMS is able to automatically send the robot to charging stations when it starts to reach low power.
    \item \textit{Robot monitoring} including the state information of the robot and the information provided by the installed sensors and cameras, etc. In case of failure or undefined behavior of the robot, the monitoring will display the warning information and a remote human operator can be contacted. 
    \item \textit{Real-time data streaming} is processed in the edge device using the real-time data recorded by the HD camera and the thermal imaging camera of the robot.
    \item \textit{DC monitoring and inspection} data flow process with the robot (in Figure~\ref{fig:ai}) including: 
    \begin{itemize}
        \item Transferring the sensor data (from various sensors) to the edge device to perform AI inference and process the images and data, such as the thermal image of a server in a rack and the DC temperature distribution and status of a server (via external status indicator LEDs on the front panel of the server);
        \item Send images and data from the robot to the RMS to generate alarms and reports.
    \end{itemize}  
    \item \textit{Security service} preventing unauthorized access to DCs through AI facial recognition processed in the robot's edge device. In this case, the spherical camera on the top of the robot is used.
\end{itemize}

Note that the identification of each rack and the servers in the rack is achieved by their predefined location in the RMS. We have proposed an equipment detection model based on the Darknet YOLO framework, similar to the work in \cite{yolo}, which uses a deep learning-based solution for real-time inspection. Darknet is an open source neural network framework written in C and CUDA (\cite{darknet}). The robot's equipment and environmental inspections are shown in Figure~\ref{fig:ee_inspection}.
\subsection{Data sources} \label{subsec:real_data}
The data layer can access data from different sources (of any type, e.g. sensor-based, reports, video data) in the hardware and service layers. A MySQL database is created in this layer and made available for use in the application layer. Multiple data representations in the database support the connection of the video stream, the information stream, the control stream and the inspection robot system. 
The data sources we use include: 
\begin{itemize}
    \item \textit{Video data} recorded by the robot's HD camera and thermal camera (using the H.264 video compression standard in MP4 (MPEG-4 Part 14) format).
    \item \textit{Inspection data} processed in the edge device; the visual inspection, thermal and acoustic inspection, and other inspection data are recorded (JSON format).
    \item \textit{Sensor data} including the environmental sensor data from the robot (JSON or XML format),
    \item \textit{Report data} including inspection reports generated by the RMS (PDF and Excel format)
    \item \textit{Robot log data} recorded while the robot is performing a task (stored in the database in XML format).   
    \item \textit{Human-robot interaction data} including human-robot interaction via tablet, camera and microphone (JSON format)
    \item \textit{VR interaction data} including VR and DT system interaction data (JSON format)     
\end{itemize}

In the DT and VR system, the Unreal Engine plugins -- Json Blueprint (\cite{JsonBlueprint}) and EasyXMLParser (\cite{EasyXMLParser}) -- are used to create and parse the JSON string and get values from XML.
\subsection{DC inspection workflow} \label{subsec:real_workflow}
\begin{figure}[h]
    \includegraphics[width=\textwidth]{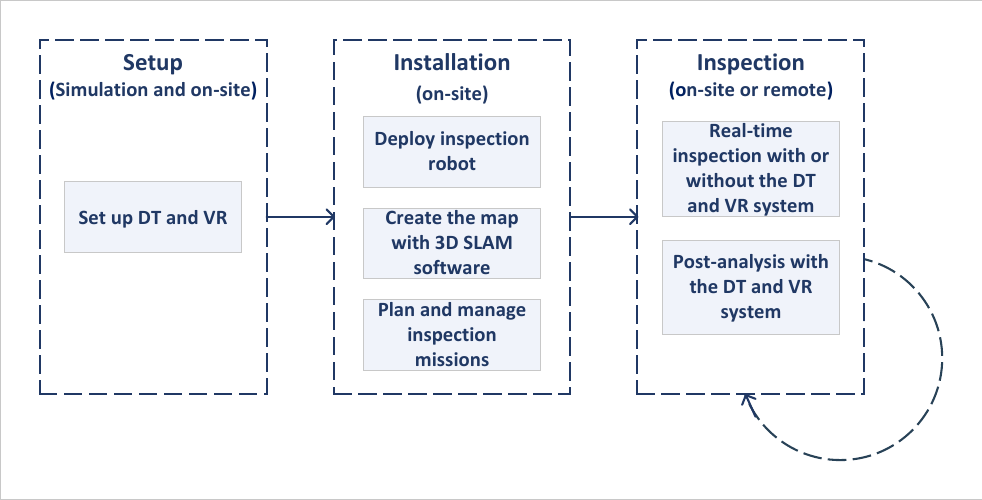}
    \caption{Flow chart presenting overview of our DT and VR in robotic inspection system in DC.}
    \label{fig:work_flow}
\end{figure}

The Pico Neo 3 Pro VR headset with installed DT and VR software, the RMS software on a computer, and an inspection robot are used in the experiments. The schematic workflow of the integrated DT and VR in the robotic inspection system is illustrated by the flow chart shown in Figure~\ref{fig:work_flow}. 

There are three phases: setup, installation, and inspection. In the setup phase, the DC and the robot are first set up in 3D using DT and VR technologies (as described in section~\ref{sec:develop}). This phase needs to be performed only once for each data center and inspection robot. 

In the installation phase, the robot is first deployed. Then the robot is used to map the DC by fusion of lidar and depth cameras. The user can use the RMS to set the start position of the robot to approximately place the robot correctly on the map. The user plans and manages inspection missions to tell the robot where to go and what to do. The missions are programmable by the user through the RMS user interface. Some functional tests of the robot's localization and navigation system are performed. The visit points of the robot and the visit path in the DC can be set in the RMS (in Figure~\ref{fig:task_plan}). The RMS synchronizes data with the robot and retrieves all site data. 

Once the installation phase is complete, inspections can be performed multiple times in the following manner, either on-site or remotely. The robot performs its inspection mission (defined in the installation step), reading inspection objects (by checking external status indicator LEDs on the front panel of the server), scanning the environment for anomalies, and recording results. An algorithm based on machine vision technology is developed to detect status indicators in low-light environments. In addition, the robot provides real-time information on the status of each mission to the VR software, including the temperature distribution data of all assets captured by the thermal imaging camera (and the resulting 3D heat map) and the data captured by the HD camera (see Figure~\ref{fig:ai}). We record the entire inspection process in the VR software. The inspection can be performed by the robot alone (automatic mode), or we can use the VR software to monitor and interact with the robot in real time during the inspection mission (manual control mode). In addition, post-analysis can be performed, such as displaying an inspection result report, replaying and analyzing the recorded inspection process, or analyzing the recorded environmental data.

\begin{figure}[h]
    \includegraphics[width=0.8\textwidth]{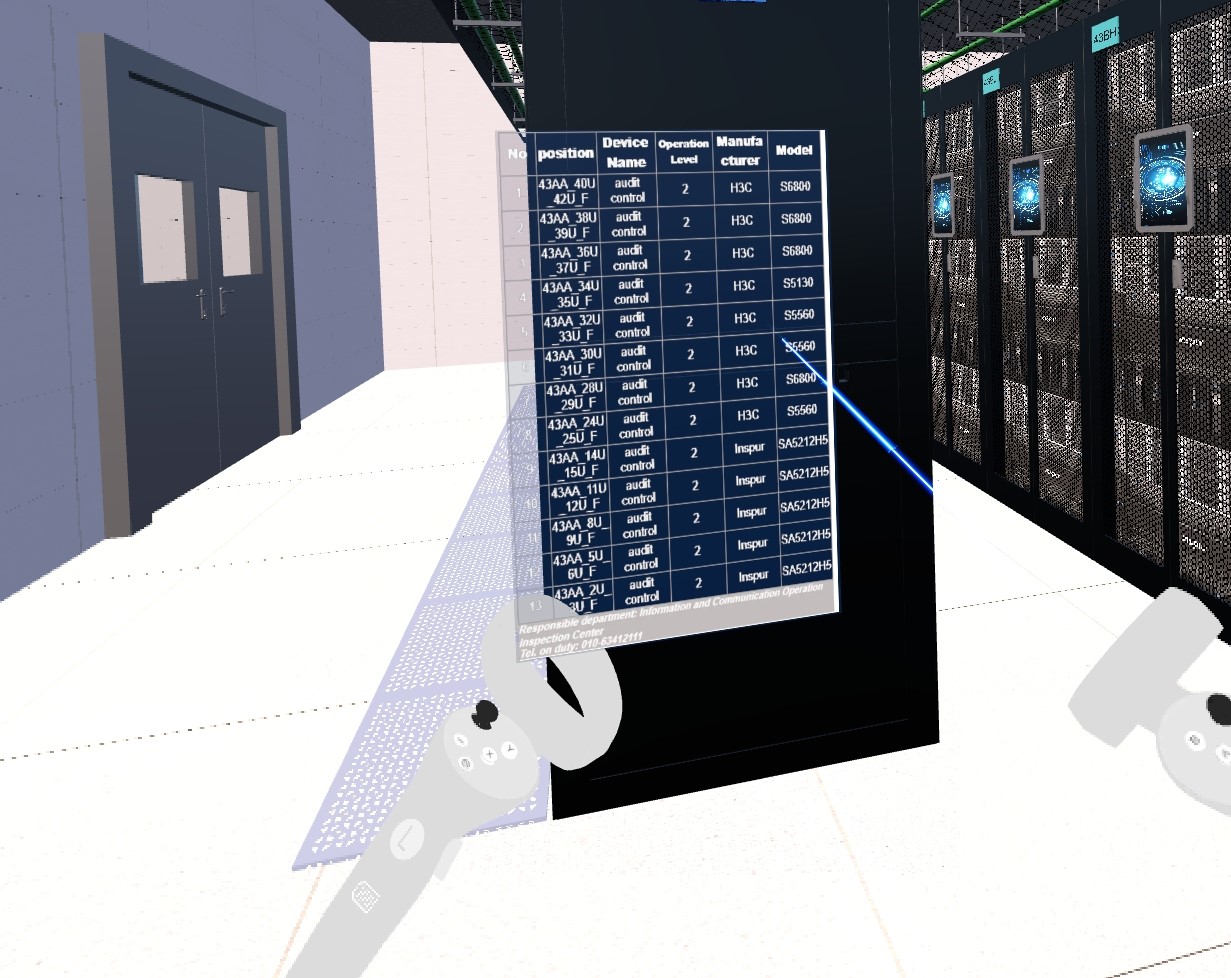}
    \caption{The knowledge graph database in VR with first-person virtual staff view.}
    \label{fig:knowledge_graph_database}
\end{figure}

We test the software from the perspective of the virtual worker and the robot. From the virtual staff perspective, we can control the virtual staff moving in the DC and interact with the virtual tablet on the front of each rack, which shows the warning or error information of the servers in the rack. By clicking on the rack, the server maintenance and specification information is displayed based on the knowledge graph database (in Figure~\ref{fig:knowledge_graph_database}). The real-time video stream, robot and data center information, and robot control are displayed by the software. From the robot perspective, we can get the warning information and set the inspection mode such as automatic and manual control. In the manual control mode, in addition to the inspection mission, we can, for example, control the lift of the robot and make some adjustments for the camera panning. 

\subsection{Results} \label{subsec:results}
We conducted a one-month test in the DC, performing daily inspections and completing a total of 30 inspection missions. As shown in Figure~\ref{fig:task_plan}, this is a typical inspection mission with 114 inspection points and an average inspection time of about 70 minutes.

By manually reviewing the recorded video captured by the robot, the accuracy rate of the automatic inspection by the robot can reach 99\% (by checking the external status indicator LEDs on the front panel of the server). The information is continuously updated in the VR and DT system. This greatly reduces the workload of DC personnel.

The system installation work needs to be done only once to create the map and DT environment, including setting up the automatic charging station and doors. For inspection, we only need to start the software, start the RMS and robot, and set the inspection mission plan. The DC staff can either start the DT and VR for real-time inspection, or do the monitoring after the automatic inspection by the robot and check the DT and VR system.

\begin{figure}
    \includegraphics[width=\textwidth]{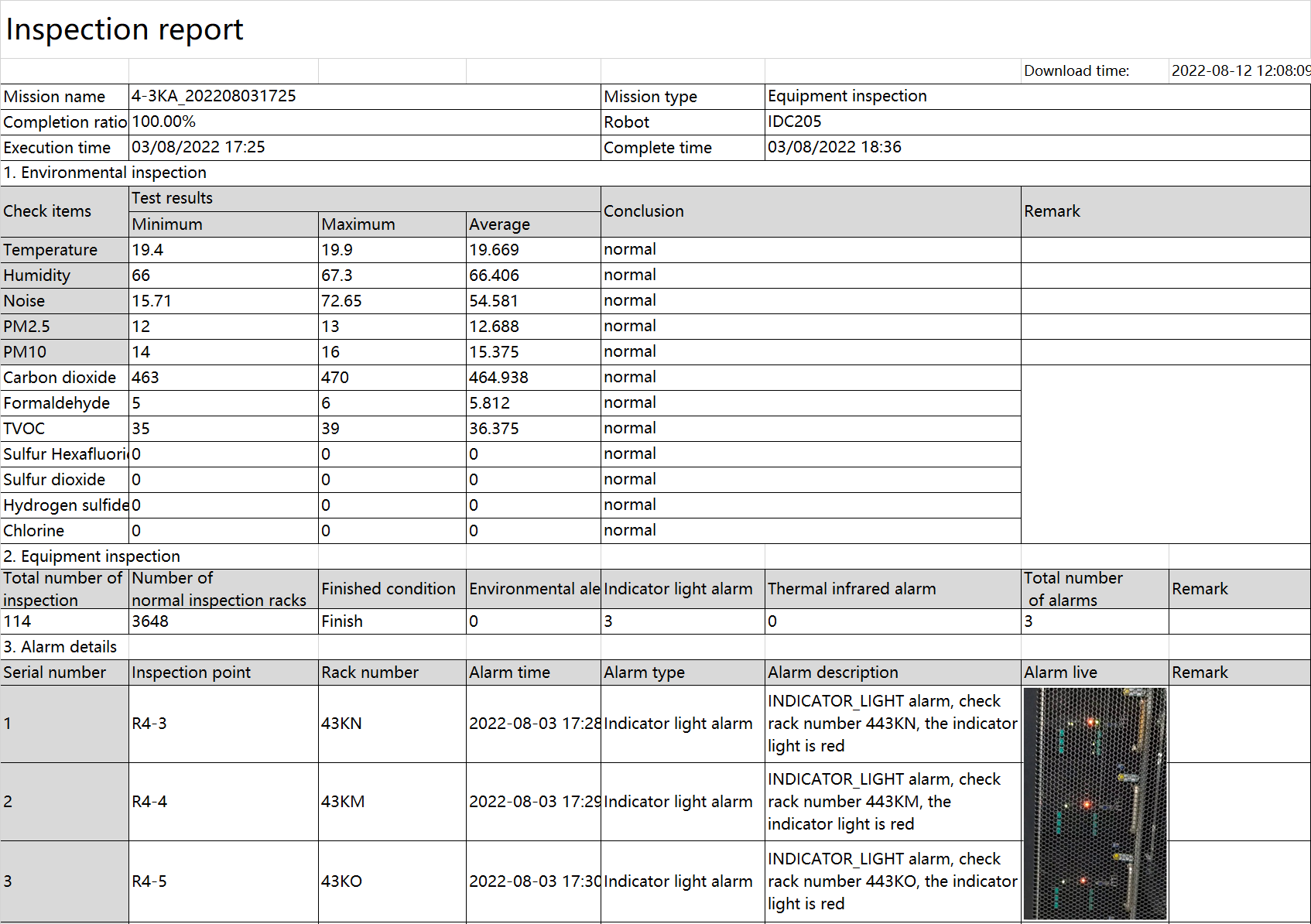}
    \caption{The inspection report.}
    \label{fig:inspection_report}
\end{figure}

Figure~\ref{fig:inspection_report} shows the captured inspection results and identified anomalies report of a typical inspection mission. Such a report consists of three parts: environmental inspection, equipment inspection, and detailed alarm information. The captured images of alarms are also included in the report. Inspection reports are saved directly in PDF and Excel formats.

\section{Conclusions}\label{sec:concl}

In this paper, we introduce a novel concept to realize a real-time immersive automated system that combines digital twin with virtual reality, since automated systems are widely used in practice. In addition, the digital twin models not only enable testing before they occur in the real systems, but also enable predictive analysis based on the digital twin data. In addition, the combination of the digital twin with virtual reality helps human personnel to interact with a robot remotely in both the physical and digital worlds. This concept will be implemented and validated to perform monitoring and inspection in a real data center to reduce downtime and ensure reliable and continuous data center services. We chose the data center application because of the rapidly increasing demand for data center services, especially during the COVID-19 pandemic. Also, our concept is considered an innovative approach in data centers that has not been considered in the literature. The validation shows that the use of the combined methodology is feasible. The system we developed has several advantages. First, the use of mobile inspection robots enables the automation of the inspection (autonomous driving and data collection via sensors/cameras), which improves productivity while reducing human workload. Thermal image analysis and acoustic analysis are automatically generated based on the data collected by the robot. Second, reliable information can be obtained through advanced sensor technology and AI-based algorithms that avoid human error and ensure the accuracy of identification. In our one-month test in a real DC, the accuracy rate can reach 99\%. Third, the human staff can work remotely using digital twin and virtual reality to control the robot to carry out real-time inspection. By using digital twin and virtual reality, the effect of human-robot collaboration can be achieved.

In the future, our concept can be applied to other automated applications, such as smart framing, power plants, and industrial processing areas (each with a different type of robot). Data analysis in digital twin data centers for fault prediction, predictive maintenance, and energy efficiency will be research topics in the near future. Validation in the near future will include user feedback, such as the usefulness and ease of interaction with the virtual reality interface. Our digital twin and virtual reality system in data centers is a promising area with high potential for enhancing the training of data center professionals. Therefore, offering training programs using our robotic inspection system may be an issue in the future. 








\bibliography{bibliography}


\begin{thebibliography}{77}
\ifx \bisbn   \undefined \def \bisbn  #1{ISBN #1}\fi
\ifx \binits  \undefined \def \binits#1{#1}\fi
\ifx \bauthor  \undefined \def \bauthor#1{#1}\fi
\ifx \batitle  \undefined \def \batitle#1{#1}\fi
\ifx \bjtitle  \undefined \def \bjtitle#1{#1}\fi
\ifx \bvolume  \undefined \def \bvolume#1{\textbf{#1}}\fi
\ifx \byear  \undefined \def \byear#1{#1}\fi
\ifx \bissue  \undefined \def \bissue#1{#1}\fi
\ifx \bfpage  \undefined \def \bfpage#1{#1}\fi
\ifx \blpage  \undefined \def \blpage #1{#1}\fi
\ifx \burl  \undefined \def \burl#1{\textsf{#1}}\fi
\ifx \doiurl  \undefined \def \doiurl#1{\url{https://doi.org/#1}}\fi
\ifx \betal  \undefined \def \betal{\textit{et al.}}\fi
\ifx \binstitute  \undefined \def \binstitute#1{#1}\fi
\ifx \binstitutionaled  \undefined \def \binstitutionaled#1{#1}\fi
\ifx \bctitle  \undefined \def \bctitle#1{#1}\fi
\ifx \beditor  \undefined \def \beditor#1{#1}\fi
\ifx \bpublisher  \undefined \def \bpublisher#1{#1}\fi
\ifx \bbtitle  \undefined \def \bbtitle#1{#1}\fi
\ifx \bedition  \undefined \def \bedition#1{#1}\fi
\ifx \bseriesno  \undefined \def \bseriesno#1{#1}\fi
\ifx \blocation  \undefined \def \blocation#1{#1}\fi
\ifx \bsertitle  \undefined \def \bsertitle#1{#1}\fi
\ifx \bsnm \undefined \def \bsnm#1{#1}\fi
\ifx \bsuffix \undefined \def \bsuffix#1{#1}\fi
\ifx \bparticle \undefined \def \bparticle#1{#1}\fi
\ifx \barticle \undefined \def \barticle#1{#1}\fi
\bibcommenthead
\ifx \bconfdate \undefined \def \bconfdate #1{#1}\fi
\ifx \botherref \undefined \def \botherref #1{#1}\fi
\ifx \url \undefined \def \url#1{\textsf{#1}}\fi
\ifx \bchapter \undefined \def \bchapter#1{#1}\fi
\ifx \bbook \undefined \def \bbook#1{#1}\fi
\ifx \bcomment \undefined \def \bcomment#1{#1}\fi
\ifx \oauthor \undefined \def \oauthor#1{#1}\fi
\ifx \citeauthoryear \undefined \def \citeauthoryear#1{#1}\fi
\ifx \endbibitem  \undefined \def \endbibitem {}\fi
\ifx \bconflocation  \undefined \def \bconflocation#1{#1}\fi
\ifx \arxivurl  \undefined \def \arxivurl#1{\textsf{#1}}\fi
\csname PreBibitemsHook\endcsname

\bibitem{rotstatistics}
\begin{botherref}
\oauthor{\bsnm{{International Federation of Robotics}}}:
World Robotics 2021 -- Service Robots report released
(2021).
\url{https://ifr.org/ifr-press-releases/news/service-robots-hit-double-digit-growth-worldwide}
\end{botherref}
\endbibitem

\bibitem{defrobot}
\begin{botherref}
\oauthor{\bsnm{{International Federation of Robotics}}}:
Definition of service robots
(2022).
\url{http://www.ifr.org/service-robots/}
\end{botherref}
\endbibitem

\bibitem{azadeh2019robotized}
\begin{barticle}
\bauthor{\bsnm{Azadeh}, \binits{K.}},
\bauthor{\bsnm{De~Koster}, \binits{R.}},
\bauthor{\bsnm{Roy}, \binits{D.}}:
\batitle{Robotized and automated warehouse systems: Review and recent developments}.
\bjtitle{Transportation Science}
\bvolume{53}(\bissue{4}),
\bfpage{917}--\blpage{945}
(\byear{2019}).
\doiurl{10.1287/trsc.2018.0873}
\end{barticle}
\endbibitem

\bibitem{xie2021introducing}
\begin{barticle}
\bauthor{\bsnm{Xie}, \binits{L.}},
\bauthor{\bsnm{Thieme}, \binits{N.}},
\bauthor{\bsnm{Krenzler}, \binits{R.}},
\bauthor{\bsnm{Li}, \binits{H.}}:
\batitle{Introducing split orders and optimizing operational policies in robotic mobile fulfillment systems}.
\bjtitle{European Journal of Operational Research}
\bvolume{288}(\bissue{1}),
\bfpage{80}--\blpage{97}
(\byear{2021}).
\doiurl{10.1016/j.ejor.2020.05.032}
\end{barticle}
\endbibitem

\bibitem{xie2022formulating}
\begin{barticle}
\bauthor{\bsnm{Xie}, \binits{L.}},
\bauthor{\bsnm{Li}, \binits{H.}},
\bauthor{\bsnm{Luttmann}, \binits{L.}}:
\batitle{Formulating and solving integrated order batching and routing in multi-depot {AGV}-assisted mixed-shelves warehouses}.
\bjtitle{European Journal of Operational Research}
(\byear{2022}).
\doiurl{10.1016/j.ejor.2022.08.047}
\end{barticle}
\endbibitem

\bibitem{merschformann2017rawsim}
\begin{barticle}
\bauthor{\bsnm{Merschformann}, \binits{M.}},
\bauthor{\bsnm{Xie}, \binits{L.}},
\bauthor{\bsnm{Li}, \binits{H.}}:
\batitle{{RAWSim-O}: A simulation framework for robotic mobile fulfillment systems}.
\bjtitle{Logistics Research}
(\byear{2018}).
\doiurl{10.23773/2018_8}
\end{barticle}
\endbibitem

\bibitem{van2021resilient}
\begin{barticle}
\bauthor{\bparticle{Van~der} \bsnm{Aalst}, \binits{W.M.}},
\bauthor{\bsnm{Hinz}, \binits{O.}},
\bauthor{\bsnm{Weinhardt}, \binits{C.}}:
\batitle{Resilient digital twins}.
\bjtitle{Business \& Information Systems Engineering}
\bvolume{22},
\bfpage{1}--\blpage{5}
(\byear{2021}).
\doiurl{10.1007/s12599-021-00721-z}
\end{barticle}
\endbibitem

\bibitem{lim2020state}
\begin{barticle}
\bauthor{\bsnm{Lim}, \binits{K.Y.H.}},
\bauthor{\bsnm{Zheng}, \binits{P.}},
\bauthor{\bsnm{Chen}, \binits{C.-H.}}:
\batitle{A state-of-the-art survey of digital twin: Techniques, engineering product lifecycle management and business innovation perspectives}.
\bjtitle{Journal of Intelligent Manufacturing}
\bvolume{31}(\bissue{6}),
\bfpage{1313}--\blpage{1337}
(\byear{2020}).
\doiurl{10.1007/s10845-019-01512-w}
\end{barticle}
\endbibitem

\bibitem{dietz2020digital}
\begin{barticle}
\bauthor{\bsnm{Dietz}, \binits{M.}},
\bauthor{\bsnm{Pernul}, \binits{G.}}:
\batitle{Digital twin: Empowering enterprises towards a system-of-systems approach}.
\bjtitle{Business \& Information Systems Engineering}
\bvolume{62}(\bissue{2}),
\bfpage{179}--\blpage{184}
(\byear{2020}).
\doiurl{10.1007/s12599-019-00624-0}
\end{barticle}
\endbibitem

\bibitem{xie2018simulation}
\begin{barticle}
\bauthor{\bsnm{Xie}, \binits{L.}},
\bauthor{\bsnm{Li}, \binits{H.}},
\bauthor{\bsnm{Thieme}, \binits{N.}}:
\batitle{From simulation to real-world robotic mobile fulfillment systems}.
\bjtitle{Logistics Research}
(\byear{2019}).
\doiurl{10.23773/2019_9}
\end{barticle}
\endbibitem

\bibitem{barosan2020development}
\begin{bchapter}
\bauthor{\bsnm{Barosan}, \binits{I.}},
\bauthor{\bsnm{Basmenj}, \binits{A.A.}},
\bauthor{\bsnm{Chouhan}, \binits{S.G.}},
\bauthor{\bsnm{Manrique}, \binits{D.}}:
\bctitle{Development of a virtual simulation environment and a digital twin of an autonomous driving truck for a distribution center}.
In: \bbtitle{European Conference on Software Architecture},
pp. \bfpage{542}--\blpage{557}
(\byear{2020}).
\doiurl{10.1007/978-3-030-59155-7_39}.
\bcomment{Springer}
\end{bchapter}
\endbibitem

\bibitem{ibrahim2022overview}
\begin{barticle}
\bauthor{\bsnm{Ibrahim}, \binits{M.}},
\bauthor{\bsnm{Rass{\~o}lkin}, \binits{A.}},
\bauthor{\bsnm{Vaimann}, \binits{T.}},
\bauthor{\bsnm{Kallaste}, \binits{A.}}:
\batitle{Overview on digital twin for autonomous electrical vehicles propulsion drive system}.
\bjtitle{Sustainability}
\bvolume{14}(\bissue{2}),
\bfpage{601}
(\byear{2022}).
\doiurl{10.3390/su14020601}
\end{barticle}
\endbibitem

\bibitem{niaz2021autonomous}
\begin{bchapter}
\bauthor{\bsnm{Niaz}, \binits{A.}},
\bauthor{\bsnm{Shoukat}, \binits{M.U.}},
\bauthor{\bsnm{Jia}, \binits{Y.}},
\bauthor{\bsnm{Khan}, \binits{S.}},
\bauthor{\bsnm{Niaz}, \binits{F.}},
\bauthor{\bsnm{Raza}, \binits{M.U.}}:
\bctitle{Autonomous driving test method based on digital twin: A survey}.
In: \bbtitle{2021 International Conference on Computing, Electronic and Electrical Engineering (ICE Cube)},
pp. \bfpage{1}--\blpage{7}
(\byear{2021}).
\doiurl{10.1109/ICECube53880.2021.9628341}.
\bcomment{IEEE}
\end{bchapter}
\endbibitem

\bibitem{lumer2021towards}
\begin{bchapter}
\bauthor{\bsnm{Lumer-Klabbers}, \binits{G.}},
\bauthor{\bsnm{Hausted}, \binits{J.O.}},
\bauthor{\bsnm{Kvistgaard}, \binits{J.L.}},
\bauthor{\bsnm{Macedo}, \binits{H.D.}},
\bauthor{\bsnm{Frasheri}, \binits{M.}},
\bauthor{\bsnm{Larsen}, \binits{P.G.}}:
\bctitle{Towards a digital twin framework for autonomous robots}.
In: \bbtitle{2021 IEEE 45th Annual Computers, Software, and Applications Conference (COMPSAC)},
pp. \bfpage{1254}--\blpage{1259}
(\byear{2021}).
\doiurl{10.1109/COMPSAC51774.2021.00174}.
\bcomment{IEEE}
\end{bchapter}
\endbibitem

\bibitem{almeaibed2021digital}
\begin{barticle}
\bauthor{\bsnm{Almeaibed}, \binits{S.}},
\bauthor{\bsnm{Al-Rubaye}, \binits{S.}},
\bauthor{\bsnm{Tsourdos}, \binits{A.}},
\bauthor{\bsnm{Avdelidis}, \binits{N.P.}}:
\batitle{Digital twin analysis to promote safety and security in autonomous vehicles}.
\bjtitle{IEEE Communications Standards Magazine}
\bvolume{5}(\bissue{1}),
\bfpage{40}--\blpage{46}
(\byear{2021}).
\doiurl{10.1109/MCOMSTD.011.2100004}
\end{barticle}
\endbibitem

\bibitem{stkaczek2021digital}
\begin{barticle}
\bauthor{\bsnm{St{\k{a}}czek}, \binits{P.}},
\bauthor{\bsnm{Pizo{\'n}}, \binits{J.}},
\bauthor{\bsnm{Danilczuk}, \binits{W.}},
\bauthor{\bsnm{Gola}, \binits{A.}}:
\batitle{A digital twin approach for the improvement of an autonomous mobile robots ({AMR}’s) operating environment--a case study}.
\bjtitle{Sensors}
\bvolume{21}(\bissue{23}),
\bfpage{7830}
(\byear{2021}).
\doiurl{10.3390/s21237830}
\end{barticle}
\endbibitem

\bibitem{cimino2019review}
\begin{barticle}
\bauthor{\bsnm{Cimino}, \binits{C.}},
\bauthor{\bsnm{Negri}, \binits{E.}},
\bauthor{\bsnm{Fumagalli}, \binits{L.}}:
\batitle{Review of digital twin applications in manufacturing}.
\bjtitle{Computers in Industry}
\bvolume{113},
\bfpage{103}--\blpage{130}
(\byear{2019}).
\doiurl{10.1016/j.compind.2019.103130}
\end{barticle}
\endbibitem

\bibitem{alatise2020review}
\begin{barticle}
\bauthor{\bsnm{Alatise}, \binits{M.B.}},
\bauthor{\bsnm{Hancke}, \binits{G.P.}}:
\batitle{A review on challenges of autonomous mobile robot and sensor fusion methods}.
\bjtitle{IEEE Access}
\bvolume{8},
\bfpage{39830}--\blpage{39846}
(\byear{2020}).
\doiurl{10.1109/ACCESS.2020.2975643}
\end{barticle}
\endbibitem

\bibitem{wohlgenannt2020virtual}
\begin{barticle}
\bauthor{\bsnm{Wohlgenannt}, \binits{I.}},
\bauthor{\bsnm{Simons}, \binits{A.}},
\bauthor{\bsnm{Stieglitz}, \binits{S.}}:
\batitle{Virtual reality}.
\bjtitle{Business \& Information Systems Engineering}
\bvolume{62}(\bissue{5}),
\bfpage{455}--\blpage{461}
(\byear{2020}).
\doiurl{10.1007/s12599-020-00658-9}
\end{barticle}
\endbibitem

\bibitem{roldan2019multi}
\begin{bchapter}
\bauthor{\bsnm{Rold{\'a}n}, \binits{J.J.}},
\bauthor{\bsnm{Pe{\~n}a-Tapia}, \binits{E.}},
\bauthor{\bsnm{Garz{\'o}n-Ramos}, \binits{D.}},
\bauthor{\bparticle{de} \bsnm{Le{\'o}n}, \binits{J.}},
\bauthor{\bsnm{Garz{\'o}n}, \binits{M.}},
\bauthor{\bparticle{del} \bsnm{Cerro}, \binits{J.}},
\bauthor{\bsnm{Barrientos}, \binits{A.}}:
\bctitle{Multi-robot systems, virtual reality and {ROS}: Developing a new generation of operator interfaces}.
In: \beditor{\bsnm{Koubaa}, \binits{A.}} (ed.)
\bbtitle{Robot Operating System (ROS)}
vol. \bseriesno{778},
pp. \bfpage{29}--\blpage{64}.
\bpublisher{Springer},
\blocation{Cham}
(\byear{2019}).
\doiurl{10.1007/978-3-319-91590-6_2}
\end{bchapter}
\endbibitem

\bibitem{oyekan2019effectiveness}
\begin{barticle}
\bauthor{\bsnm{Oyekan}, \binits{J.O.}},
\bauthor{\bsnm{Hutabarat}, \binits{W.}},
\bauthor{\bsnm{Tiwari}, \binits{A.}},
\bauthor{\bsnm{Grech}, \binits{R.}},
\bauthor{\bsnm{Aung}, \binits{M.H.}},
\bauthor{\bsnm{Mariani}, \binits{M.P.}},
\bauthor{\bsnm{L{\'o}pez-D{\'a}valos}, \binits{L.}},
\bauthor{\bsnm{Ricaud}, \binits{T.}},
\bauthor{\bsnm{Singh}, \binits{S.}},
\bauthor{\bsnm{Dupuis}, \binits{C.}}:
\batitle{The effectiveness of virtual environments in developing collaborative strategies between industrial robots and humans}.
\bjtitle{Robotics and Computer-Integrated Manufacturing}
\bvolume{55},
\bfpage{41}--\blpage{54}
(\byear{2019}).
\doiurl{10.1016/j.rcim.2018.07.006}
\end{barticle}
\endbibitem

\bibitem{kot2018application}
\begin{barticle}
\bauthor{\bsnm{Kot}, \binits{T.}},
\bauthor{\bsnm{Nov{\'a}k}, \binits{P.}}:
\batitle{Application of virtual reality in teleoperation of the military mobile robotic system {TAROS}}.
\bjtitle{International Journal of Advanced Robotic Systems}
\bvolume{15}(\bissue{1}),
\bfpage{1729881417751545}
(\byear{2018}).
\doiurl{10.1177/1729881417751545}
\end{barticle}
\endbibitem

\bibitem{li2022towards}
\begin{bchapter}
\bauthor{\bsnm{Li}, \binits{K.}},
\bauthor{\bsnm{Bacher}, \binits{R.}},
\bauthor{\bsnm{Leemans}, \binits{W.}},
\bauthor{\bsnm{Steinicke}, \binits{F.}}:
\bctitle{Towards robust exocentric mobile robot tele-operation in mixed reality}.
In: \bbtitle{5th International Workshop on Virtual, Augmented, and Mixed Reality for HRI}
(\byear{2022}).
\burl{https://openreview.net/forum?id=HYIes841hJc}
\end{bchapter}
\endbibitem

\bibitem{wonsick2020systematic}
\begin{barticle}
\bauthor{\bsnm{Wonsick}, \binits{M.}},
\bauthor{\bsnm{Padir}, \binits{T.}}:
\batitle{A systematic review of virtual reality interfaces for controlling and interacting with robots}.
\bjtitle{Applied Sciences}
\bvolume{10}(\bissue{24}),
\bfpage{9051}
(\byear{2020}).
\doiurl{10.3390/app10249051}
\end{barticle}
\endbibitem

\bibitem{havard2019digital}
\begin{barticle}
\bauthor{\bsnm{Havard}, \binits{V.}},
\bauthor{\bsnm{Jeanne}, \binits{B.}},
\bauthor{\bsnm{Lacomblez}, \binits{M.}},
\bauthor{\bsnm{Baudry}, \binits{D.}}:
\batitle{Digital twin and virtual reality: A co-simulation environment for design and assessment of industrial workstations}.
\bjtitle{Production \& Manufacturing Research}
\bvolume{7}(\bissue{1}),
\bfpage{472}--\blpage{489}
(\byear{2019}).
\doiurl{10.1080/21693277.2019.1660283}
\end{barticle}
\endbibitem

\bibitem{perez2020digital}
\begin{barticle}
\bauthor{\bsnm{P{\'e}rez}, \binits{L.}},
\bauthor{\bsnm{Rodr{\'\i}guez-Jim{\'e}nez}, \binits{S.}},
\bauthor{\bsnm{Rodr{\'\i}guez}, \binits{N.}},
\bauthor{\bsnm{Usamentiaga}, \binits{R.}},
\bauthor{\bsnm{Garcia}, \binits{D.F.}}:
\batitle{Digital twin and virtual reality based methodology for multi-robot manufacturing cell commissioning}.
\bjtitle{Applied Sciences}
\bvolume{10}(\bissue{10}),
\bfpage{3633}
(\byear{2020}).
\doiurl{10.3390/app10103633}
\end{barticle}
\endbibitem

\bibitem{burghardt2020programming}
\begin{barticle}
\bauthor{\bsnm{Burghardt}, \binits{A.}},
\bauthor{\bsnm{Szybicki}, \binits{D.}},
\bauthor{\bsnm{Gierlak}, \binits{P.}},
\bauthor{\bsnm{Kurc}, \binits{K.}},
\bauthor{\bsnm{Pietru{\'s}}, \binits{P.}},
\bauthor{\bsnm{Cygan}, \binits{R.}}:
\batitle{Programming of industrial robots using virtual reality and digital twins}.
\bjtitle{Applied Sciences}
\bvolume{10}(\bissue{2}),
\bfpage{486}
(\byear{2020}).
\doiurl{10.3390/app10020486}
\end{barticle}
\endbibitem

\bibitem{kuts2019digital}
\begin{barticle}
\bauthor{\bsnm{Kuts}, \binits{V.}},
\bauthor{\bsnm{Otto}, \binits{T.}},
\bauthor{\bsnm{T{\"a}hemaa}, \binits{T.}},
\bauthor{\bsnm{Bondarenko}, \binits{Y.}}:
\batitle{Digital twin based synchronised control and simulation of the industrial robotic cell using virtual reality}.
\bjtitle{Journal of Machine Engineering}
\bvolume{19}(\bissue{1}),
\bfpage{128}--\blpage{145}
(\byear{2019}).
\doiurl{10.5604/01.3001.0013.0464}
\end{barticle}
\endbibitem

\bibitem{rocca2020integrating}
\begin{barticle}
\bauthor{\bsnm{Rocca}, \binits{R.}},
\bauthor{\bsnm{Rosa}, \binits{P.}},
\bauthor{\bsnm{Sassanelli}, \binits{C.}},
\bauthor{\bsnm{Fumagalli}, \binits{L.}},
\bauthor{\bsnm{Terzi}, \binits{S.}}:
\batitle{Integrating virtual reality and digital twin in circular economy practices: A laboratory application case}.
\bjtitle{Sustainability}
\bvolume{12}(\bissue{6}),
\bfpage{2286}
(\byear{2020}).
\doiurl{10.3390/su12062286}
\end{barticle}
\endbibitem

\bibitem{williams2020augmented}
\begin{barticle}
\bauthor{\bsnm{Williams}, \binits{R.}},
\bauthor{\bsnm{Erkoyuncu}, \binits{J.A.}},
\bauthor{\bsnm{Masood}, \binits{T.}},
\bauthor{\bsnm{Vrabic}, \binits{R.}}:
\batitle{Augmented reality assisted calibration of digital twins of mobile robots}.
\bjtitle{IFAC-PapersOnLine}
\bvolume{53}(\bissue{3}),
\bfpage{203}--\blpage{208}
(\byear{2020}).
\doiurl{10.1016/j.ifacol.2020.11.033}
\end{barticle}
\endbibitem

\bibitem{kaarlela2020digital}
\begin{bchapter}
\bauthor{\bsnm{Kaarlela}, \binits{T.}},
\bauthor{\bsnm{Piesk{\"a}}, \binits{S.}},
\bauthor{\bsnm{Pitk{\"a}aho}, \binits{T.}}:
\bctitle{Digital twin and virtual reality for safety training}.
In: \bbtitle{2020 11th IEEE International Conference on Cognitive Infocommunications (CogInfoCom)},
pp. \bfpage{115}--\blpage{120}
(\byear{2020}).
\doiurl{10.1109/CogInfoCom50765.2020.9237812}
\end{bchapter}
\endbibitem

\bibitem{bieser2018role}
\begin{bchapter}
\bauthor{\bsnm{Bieser}, \binits{J.}},
\bauthor{\bsnm{Kunbaz}, \binits{M.}}:
\bctitle{The role of facility maintenance in data centres: A case study}.
In: \bbtitle{European Facility Management Conference (EFMC 2018)}
(\byear{2018})
\end{bchapter}
\endbibitem

\bibitem{datacenter}
\begin{botherref}
\oauthor{\bsnm{Haranas}, \binits{M.}}:
{AWS}, {G}oogle, {M}icrosoft are taking over the data center market
(2021).
\url{https://www.crn.com/news/data-center/aws-google-microsoft-are-taking-over-the-data-center}
\end{botherref}
\endbibitem

\bibitem{ponemon}
\begin{botherref}
\oauthor{\bsnm{{Ponemon Institute}}}:
Cost of data center outages
(2016).
\url{{https://planetaklimata.com.ua/instr/Liebert_Hiross/Cost_of_Data_Center_Outages_2016_Eng.pdf}}
\end{botherref}
\endbibitem

\bibitem{gill2011understanding}
\begin{bchapter}
\bauthor{\bsnm{Gill}, \binits{P.}},
\bauthor{\bsnm{Jain}, \binits{N.}},
\bauthor{\bsnm{Nagappan}, \binits{N.}}:
\bctitle{Understanding network failures in data centers: Measurement, analysis, and implications}.
In: \bbtitle{Proceedings of the ACM SIGCOMM 2011 Conference},
pp. \bfpage{350}--\blpage{361}
(\byear{2011}).
\doiurl{10.1145/2018436.2018477}
\end{bchapter}
\endbibitem

\bibitem{guenter2011managing}
\begin{bchapter}
\bauthor{\bsnm{Guenter}, \binits{B.}},
\bauthor{\bsnm{Jain}, \binits{N.}},
\bauthor{\bsnm{Williams}, \binits{C.}}:
\bctitle{Managing cost, performance, and reliability tradeoffs for energy-aware server provisioning}.
In: \bbtitle{2011 Proceedings IEEE INFOCOM},
pp. \bfpage{1332}--\blpage{1340}
(\byear{2011}).
\doiurl{10.1109/infcom.2011.5934917}.
\bcomment{IEEE}
\end{bchapter}
\endbibitem

\bibitem{ram2014modeling}
\begin{barticle}
\bauthor{\bsnm{Ram}, \binits{M.}},
\bauthor{\bsnm{Singh}, \binits{V.}}:
\batitle{Modeling and availability analysis of internet data center with various maintenance policies}.
\bjtitle{International Journal of Engineering}
\bvolume{27}(\bissue{4}),
\bfpage{599}--\blpage{608}
(\byear{2014}).
\doiurl{10.5829/idosi.ije.2014.27.04a.10}
\end{barticle}
\endbibitem

\bibitem{wang2009towards}
\begin{bchapter}
\bauthor{\bsnm{Wang}, \binits{L.}},
\bauthor{\bparticle{von} \bsnm{Laszewski}, \binits{G.}},
\bauthor{\bsnm{Dayal}, \binits{J.}},
\bauthor{\bsnm{He}, \binits{X.}},
\bauthor{\bsnm{Younge}, \binits{A.J.}},
\bauthor{\bsnm{Furlani}, \binits{T.R.}}:
\bctitle{Towards thermal aware workload scheduling in a data center}.
In: \bbtitle{10th International Symposium on Pervasive Systems, Algorithms, and Networks},
pp. \bfpage{116}--\blpage{122}
(\byear{2009}).
\doiurl{10.1109/i-span.2009.22}.
\bcomment{IEEE}
\end{bchapter}
\endbibitem

\bibitem{bayati2016managing}
\begin{bchapter}
\bauthor{\bsnm{Bayati}, \binits{M.}}:
\bctitle{Managing energy consumption and quality of service in data centers}.
In: \bbtitle{5th International Conference on Smart Cities and Green ICT Systems (SMARTGREENS)},
pp. \bfpage{293}--\blpage{301}
(\byear{2016}).
\doiurl{10.5220/0005791802930301}.
\bcomment{IEEE}
\end{bchapter}
\endbibitem

\bibitem{chen2013optimization}
\begin{bchapter}
\bauthor{\bsnm{Chen}, \binits{S.}},
\bauthor{\bsnm{Hu}, \binits{Y.}},
\bauthor{\bsnm{Peng}, \binits{L.}}:
\bctitle{Optimization of electricity and server maintenance costs in hybrid cooling data centers}.
In: \bbtitle{2013 IEEE Sixth International Conference on Cloud Computing},
pp. \bfpage{526}--\blpage{533}
(\byear{2013}).
\doiurl{10.1109/cloud.2013.104}.
\bcomment{IEEE}
\end{bchapter}
\endbibitem

\bibitem{cho2015development}
\begin{barticle}
\bauthor{\bsnm{Cho}, \binits{J.}},
\bauthor{\bsnm{Yang}, \binits{J.}},
\bauthor{\bsnm{Lee}, \binits{C.}},
\bauthor{\bsnm{Lee}, \binits{J.}}:
\batitle{Development of an energy evaluation and design tool for dedicated cooling systems of data centers: Sensing data center cooling energy efficiency}.
\bjtitle{Energy and Buildings}
\bvolume{96},
\bfpage{357}--\blpage{372}
(\byear{2015}).
\doiurl{10.1016/j.enbuild.2015.03.040}
\end{barticle}
\endbibitem

\bibitem{li2016data}
\begin{barticle}
\bauthor{\bsnm{Li}, \binits{L.}},
\bauthor{\bsnm{Zheng}, \binits{W.}},
\bauthor{\bsnm{Wang}, \binits{X.}},
\bauthor{\bsnm{Wang}, \binits{X.}}:
\batitle{Data center power minimization with placement optimization of liquid-cooled servers and free air cooling}.
\bjtitle{Sustainable Computing: Informatics and Systems}
\bvolume{11},
\bfpage{3}--\blpage{15}
(\byear{2016}).
\doiurl{10.1016/j.suscom.2016.02.001}
\end{barticle}
\endbibitem

\bibitem{ignore}
\begin{botherref}
\oauthor{\bsnm{Clark}, \binits{J.}}:
You can’t afford to ignore data center maintenance
(2012).
\url{https://www.datacenterjournal.com/it/you-cant-afford-to-ignore-data-center-maintenance/}
\end{botherref}
\endbibitem

\bibitem{bieser2019assessing}
\begin{bchapter}
\bauthor{\bsnm{Bieser}, \binits{J.}},
\bauthor{\bsnm{Menzel}, \binits{K.}}:
\bctitle{Assessing facility maintenance models for data centres: Status and deficits of current facility management and maintenance concepts}.
In: \bbtitle{Applied Mechanics and Materials},
vol. \bseriesno{887},
pp. \bfpage{255}--\blpage{263}
(\byear{2019}).
\doiurl{10.4028/www.scientific.net/AMM.887.255}.
\bcomment{Trans Tech Publ}
\end{bchapter}
\endbibitem

\bibitem{bieser2018assessing}
\begin{botherref}
\oauthor{\bsnm{Bieser}, \binits{J.}},
\oauthor{\bsnm{Menzel}, \binits{K.}},
\oauthor{\bsnm{Hoffmann}, \binits{K.}}:
Assessing a facility maintenance model of data centers: A methodology for advanced maintenance management for data centers.
ICCCBE, Finland
(2018)
\end{botherref}
\endbibitem

\bibitem{abadi2020data}
\begin{barticle}
\bauthor{\bsnm{Abadi}, \binits{M.F.}},
\bauthor{\bsnm{Haghighat}, \binits{F.}},
\bauthor{\bsnm{Nasiri}, \binits{F.}}:
\batitle{Data center maintenance: Applications and future research directions}.
\bjtitle{Facilities}
\bvolume{38}(\bissue{9/10}),
\bfpage{691}--\blpage{714}
(\byear{2020}).
\doiurl{10.1108/F-09-2019-0104}
\end{barticle}
\endbibitem

\bibitem{kadir2015wireless}
\begin{bchapter}
\bauthor{\bsnm{Kadir}, \binits{E.A.}},
\bauthor{\bsnm{Shamsuddin}, \binits{S.M.}},
\bauthor{\bsnm{Hasan}, \binits{S.}},
\bauthor{\bsnm{Rosa}, \binits{S.L.}}:
\bctitle{Wireless monitoring for big data center server room and equipments}.
In: \bbtitle{2015 International Conference on Science in Information Technology (ICSITech)},
pp. \bfpage{187}--\blpage{191}
(\byear{2015}).
\doiurl{10.1109/icsitech.2015.7407801}.
\bcomment{IEEE}
\end{bchapter}
\endbibitem

\bibitem{hubbell2015big}
\begin{bchapter}
\bauthor{\bsnm{Hubbell}, \binits{M.}},
\bauthor{\bsnm{Moran}, \binits{A.}},
\bauthor{\bsnm{Arcand}, \binits{W.}},
\bauthor{\bsnm{Bestor}, \binits{D.}},
\bauthor{\bsnm{Bergeron}, \binits{B.}},
\bauthor{\bsnm{Byun}, \binits{C.}},
\bauthor{\bsnm{Gadepally}, \binits{V.}},
\bauthor{\bsnm{Michaleas}, \binits{P.}},
\bauthor{\bsnm{Mullen}, \binits{J.}},
\bauthor{\bsnm{Prout}, \binits{A.}}, \betal:
\bctitle{Big data strategies for data center infrastructure management using a {3D} gaming platform}.
In: \bbtitle{2015 IEEE High Performance Extreme Computing Conference (HPEC)},
pp. \bfpage{1}--\blpage{6}
(\byear{2015}).
\doiurl{10.1109/hpec.2015.7322471}.
\bcomment{IEEE}
\end{bchapter}
\endbibitem

\bibitem{levy2021emerging}
\begin{bchapter}
\bauthor{\bsnm{Levy}, \binits{M.}},
\bauthor{\bsnm{Subburaj}, \binits{A.}}:
\bctitle{Emerging trends in data center management automation}.
In: \bbtitle{2021 IEEE 11th Annual Computing and Communication Workshop and Conference (CCWC)},
pp. \bfpage{480}--\blpage{485}
(\byear{2021}).
\doiurl{10.1109/ccwc51732.2021.9375837}.
\bcomment{IEEE}
\end{bchapter}
\endbibitem

\bibitem{decker2019big}
\begin{bchapter}
\bauthor{\bparticle{Decker~de} \bsnm{Sousa}, \binits{L.}},
\bauthor{\bsnm{Giommi}, \binits{L.}},
\bauthor{\bsnm{Rossi}, \binits{T.S.}},
\bauthor{\bsnm{Viola}, \binits{F.}},
\bauthor{\bsnm{Martelli}, \binits{B.}},
\bauthor{\bsnm{Bonacorsi}, \binits{D.}}:
\bctitle{Big data analysis for predictive maintenance at the {INFN-CNAF} data center using machine learning approaches}.
In: \bbtitle{Conference of Open Innovations Association},
vol. \bseriesno{622},
pp. \bfpage{448}--\blpage{451}
(\byear{2019}).
\bcomment{FRUCT Oy}
\end{bchapter}
\endbibitem

\bibitem{gehring2021anymal}
\begin{bchapter}
\bauthor{\bsnm{Gehring}, \binits{C.}},
\bauthor{\bsnm{Fankhauser}, \binits{P.}},
\bauthor{\bsnm{Isler}, \binits{L.}},
\bauthor{\bsnm{Diethelm}, \binits{R.}},
\bauthor{\bsnm{Bachmann}, \binits{S.}},
\bauthor{\bsnm{Potz}, \binits{M.}},
\bauthor{\bsnm{Gerstenberg}, \binits{L.}},
\bauthor{\bsnm{Hutter}, \binits{M.}}:
\bctitle{{ANYmal} in the field: Solving industrial inspection of an offshore {HVDC} platform with a quadrupedal robot}.
In: \bbtitle{Field and Service Robotics},
pp. \bfpage{247}--\blpage{260}
(\byear{2021}).
\doiurl{10.3929/ethz-b-000360083}.
\bcomment{Springer}
\end{bchapter}
\endbibitem

\bibitem{alhassan2020power}
\begin{barticle}
\bauthor{\bsnm{Alhassan}, \binits{A.B.}},
\bauthor{\bsnm{Zhang}, \binits{X.}},
\bauthor{\bsnm{Shen}, \binits{H.}},
\bauthor{\bsnm{Xu}, \binits{H.}}:
\batitle{Power transmission line inspection robots: A review, trends and challenges for future research}.
\bjtitle{International Journal of Electrical Power \& Energy Systems}
\bvolume{118},
\bfpage{105862}
(\byear{2020}).
\doiurl{10.1016/j.ijepes.2020.105862}
\end{barticle}
\endbibitem

\bibitem{yu2019inspection}
\begin{bchapter}
\bauthor{\bsnm{Yu}, \binits{L.}},
\bauthor{\bsnm{Yang}, \binits{E.}},
\bauthor{\bsnm{Ren}, \binits{P.}},
\bauthor{\bsnm{Luo}, \binits{C.}},
\bauthor{\bsnm{Dobie}, \binits{G.}},
\bauthor{\bsnm{Gu}, \binits{D.}},
\bauthor{\bsnm{Yan}, \binits{X.}}:
\bctitle{Inspection robots in oil and gas industry: A review of current solutions and future trends}.
In: \bbtitle{25th International Conference on Automation and Computing (ICAC)},
pp. \bfpage{1}--\blpage{6}
(\byear{2019}).
\doiurl{10.23919/iconac.2019.8895089}.
\bcomment{IEEE}
\end{bchapter}
\endbibitem

\bibitem{zimroz2019should}
\begin{bchapter}
\bauthor{\bsnm{Zimroz}, \binits{R.}},
\bauthor{\bsnm{Hutter}, \binits{M.}},
\bauthor{\bsnm{Mistry}, \binits{M.}},
\bauthor{\bsnm{Stefaniak}, \binits{P.}},
\bauthor{\bsnm{Walas}, \binits{K.}},
\bauthor{\bsnm{Wodecki}, \binits{J.}}:
\bctitle{Why should inspection robots be used in deep underground mines?}
In: \beditor{\bsnm{Widzyk-Capehart~E.}, \binits{S.R.} \bsuffix{Hekmat~A.}} (ed.)
\bbtitle{Proceedings of the 27th International Symposium on Mine Planning and Equipment Selection--MPES 2018},
pp. \bfpage{497}--\blpage{507}
(\byear{2019}).
\doiurl{10.1007/978-3-319-99220-4_42}.
\bcomment{Springer}
\end{bchapter}
\endbibitem

\bibitem{kroll2008survey}
\begin{bchapter}
\bauthor{\bsnm{Kroll}, \binits{A.}}:
\bctitle{A survey on mobile robots for industrial inspection}.
In: \bbtitle{Proceedings of the International Conference on Intelligent Autonomous Systems IAS10},
pp. \bfpage{406}--\blpage{414}
(\byear{2008}).
\doiurl{10.3233/978-1-58603-887-8-406}
\end{bchapter}
\endbibitem

\bibitem{katrasnik2009survey}
\begin{barticle}
\bauthor{\bsnm{Katrasnik}, \binits{J.}},
\bauthor{\bsnm{Pernus}, \binits{F.}},
\bauthor{\bsnm{Likar}, \binits{B.}}:
\batitle{A survey of mobile robots for distribution power line inspection}.
\bjtitle{IEEE Transactions on Power Delivery}
\bvolume{25}(\bissue{1}),
\bfpage{485}--\blpage{493}
(\byear{2009}).
\doiurl{10.1109/tpwrd.2009.2035427}
\end{barticle}
\endbibitem

\bibitem{rosa2014towards}
\begin{bchapter}
\bauthor{\bsnm{Rosa}, \binits{S.}},
\bauthor{\bsnm{Russo}, \binits{L.O.}},
\bauthor{\bsnm{Bona}, \binits{B.}}:
\bctitle{Towards a {ROS}-based autonomous cloud robotics platform for data center monitoring}.
In: \bbtitle{Proceedings of the 2014 IEEE Emerging Technology and Factory Automation (ETFA)},
pp. \bfpage{1}--\blpage{8}
(\byear{2014}).
\doiurl{10.1109/etfa.2014.7005212}.
\bcomment{IEEE}
\end{bchapter}
\endbibitem

\bibitem{russo2016novel}
\begin{barticle}
\bauthor{\bsnm{Russo}, \binits{L.O.}},
\bauthor{\bsnm{Rosa}, \binits{S.}},
\bauthor{\bsnm{Maggiora}, \binits{M.}},
\bauthor{\bsnm{Bona}, \binits{B.}}:
\batitle{A novel cloud-based service robotics application to data center environmental monitoring}.
\bjtitle{Sensors}
\bvolume{16}(\bissue{8}),
\bfpage{1255}
(\byear{2016}).
\doiurl{10.3390/s16081255}
\end{barticle}
\endbibitem

\bibitem{terrissa2019robotics}
\begin{bchapter}
\bauthor{\bsnm{Terrissa}, \binits{L.S.}},
\bauthor{\bsnm{Ayad}, \binits{S.}},
\bauthor{\bsnm{Zerhouni}, \binits{N.}}:
\bctitle{Robotics based solution for data center e-monitoring}.
In: \bbtitle{2019 International Conference on Advanced Systems and Emergent Technologies (IC\_ASET)},
pp. \bfpage{201}--\blpage{208}
(\byear{2019}).
\doiurl{10.1109/aset.2019.8871015}.
\bcomment{IEEE}
\end{bchapter}
\endbibitem

\bibitem{Citigroup}
\begin{botherref}
\oauthor{\bsnm{{Future Facilities}}}:
Data center Digital Twin -- {A} necessity for unlocking your data center’s business potential
(2020).
\url{https://www.futurefacilities.com/the-data-center-digital-twin/}
\end{botherref}
\endbibitem

\bibitem{khan2021inspection}
\begin{barticle}
\bauthor{\bsnm{Khan}, \binits{A.}},
\bauthor{\bsnm{Mineo}, \binits{C.}},
\bauthor{\bsnm{Dobie}, \binits{G.}},
\bauthor{\bsnm{Macleod}, \binits{C.}},
\bauthor{\bsnm{Pierce}, \binits{G.}}:
\batitle{Vision guided robotic inspection for parts in manufacturing and remanufacturing industry}.
\bjtitle{Journal of Remanufacturing}
\bvolume{11}(\bissue{1}),
\bfpage{49}--\blpage{70}
(\byear{2021}).
\doiurl{10.1007/s13243-020-00091-x}
\end{barticle}
\endbibitem

\bibitem{RobotsInDCs2}
\begin{botherref}
\oauthor{\bsnm{Moss}, \binits{S.}}:
The slow rise of robots in the data center
(2021).
\url{https://www.datacenterdynamics.com/en/analysis/caves-of-steel/}
\end{botherref}
\endbibitem

\bibitem{RobotsInDCs1}
\begin{botherref}
\oauthor{\bsnm{Sagar}, \binits{R.}}:
Current state of robots at the data centers
(2021).
\url{https://analyticsindiamag.com/current-state-of-robots-at-the-data-centers/}
\end{botherref}
\endbibitem

\bibitem{seo2019using}
\begin{barticle}
\bauthor{\bsnm{Seo}, \binits{J.H.}},
\bauthor{\bsnm{Bruner}, \binits{M.}},
\bauthor{\bsnm{Payne}, \binits{A.}},
\bauthor{\bsnm{Gober}, \binits{N.}},
\bauthor{\bsnm{McMullen}, \binits{D.}}:
\batitle{Using virtual reality to enforce principles of cybersecurity}.
\bjtitle{Journal of Computational Science Education}
\bvolume{10}(\bissue{1}),
\bfpage{81}--\blpage{87}
(\byear{2019}).
\doiurl{10.22369/issn.2153-4136/10/1/13}
\end{barticle}
\endbibitem

\bibitem{VR2018book}
\begin{bbook}
\bauthor{\bsnm{Sherman}, \binits{W.R.}},
\bauthor{\bsnm{Craig}, \binits{A.B.}}:
\bbtitle{Understanding Virtual Reality: Interface, Application, and Design},
\bedition{2}nd edn.
\bpublisher{Morgan Kaufmann},
\blocation{Burlington, Massachusetts}
(\byear{2018}).
\doiurl{10.1016/C2013-0-18583-2}
\end{bbook}
\endbibitem

\bibitem{MetaHuman}
\begin{botherref}
\oauthor{\bsnm{{Unreal Engine}}}:
MetaHuman
(2022).
\url{https://www.unrealengine.com/en-US/metahuman}
\end{botherref}
\endbibitem

\bibitem{Kuts2019dt}
\begin{barticle}
\bauthor{\bsnm{Kuts}, \binits{V.}},
\bauthor{\bsnm{Otto}, \binits{T.}},
\bauthor{\bsnm{Tähemaa}, \binits{T.}},
\bauthor{\bsnm{Bondarenko}, \binits{Y.}}:
\batitle{Digital twin based synchronised control and simulation of the industrial robotic cell using virtual reality}.
\bjtitle{Journal of Machine Engineering}
\bvolume{19}(\bissue{1}),
\bfpage{128}--\blpage{144}
(\byear{2019}).
\doiurl{10.5604/01.3001.0013.0464}
\end{barticle}
\endbibitem

\bibitem{OpenXR}
\begin{botherref}
\oauthor{\bsnm{{Unreal Engine}}}:
Developing for head-mounted experiences with OpenXR
(2022).
\url{https://docs.unrealengine.com/5.0/en-US/developing-for-head-mounted-experiences-with-openxr-in-unreal-engine/}
\end{botherref}
\endbibitem

\bibitem{MultiTask2}
\begin{botherref}
\oauthor{\bsnm{{Pug Life Studio}}}:
Unreal Engine plugin -- Multi Task 2
(2022).
\url{https://www.unrealengine.com/marketplace/en-US/product/multi-task-01}
\end{botherref}
\endbibitem

\bibitem{VlcMedia}
\begin{botherref}
\oauthor{\bsnm{Helmers}, \binits{T.}}:
Unreal Engine 4 Media Framework plugin using the Video LAN Codec
(2022).
\url{https://github.com/helmers-timo/VlcMedia}
\end{botherref}
\endbibitem

\bibitem{VaRest}
\begin{botherref}
\oauthor{\bsnm{Alyamkin}, \binits{V.}}:
Unreal Engine plugin -- {VaRest}
(2022).
\url{https://www.unrealengine.com/marketplace/en-US/product/varest-plugin}
\end{botherref}
\endbibitem

\bibitem{DataRegistries}
\begin{botherref}
\oauthor{\bsnm{{Unreal Engine}}}:
Unreal Engine plugin -- {Data Registries}
(2022).
\url{https://docs.unrealengine.com/5.0/en-US/data-registries-in-unreal-engine/}
\end{botherref}
\endbibitem

\bibitem{MathInterpolation}
\begin{botherref}
\oauthor{\bsnm{{Unreal Engine}}}:
Unreal Engine blueprint API -- Math Interpolation
(2022).
\url{https://docs.unrealengine.com/5.0/en-US/BlueprintAPI/Math/Interpolation/}
\end{botherref}
\endbibitem

\bibitem{yolo}
\begin{botherref}
\oauthor{\bsnm{Redmon}, \binits{J.}},
\oauthor{\bsnm{Farhadi}, \binits{A.}}:
{YOLOv3}: An incremental improvement
(2018).
\url{https://arxiv.org/abs/1804.02767?e05802c1}
\end{botherref}
\endbibitem

\bibitem{darknet}
\begin{botherref}
\oauthor{\bsnm{Redmon}, \binits{J.}}:
Darknet: Open source neural networks in C
(2022).
\url{http://pjreddie.com/darknet/}
\end{botherref}
\endbibitem

\bibitem{JsonBlueprint}
\begin{botherref}
\oauthor{\bsnm{Shestakov}, \binits{M.}}:
Unreal Engine plugin -- Json Blueprint
(2022).
\url{https://www.unrealengine.com/marketplace/en-US/product/json-blueprint}
\end{botherref}
\endbibitem

\bibitem{EasyXMLParser}
\begin{botherref}
\oauthor{\bsnm{Ayumax}}:
Unreal Engine plugin -- EasyXMLParser
(2022).
\url{https://www.unrealengine.com/marketplace/en-US/product/easyxmlparser}
\end{botherref}
\endbibitem

\end{thebibliography}


\end{document}